\documentclass[journal]{IEEEtran}

\usepackage{cite}

\usepackage{calc,amsfonts,amssymb,amsmath,bm,url,color,theorem,graphicx,cite}
\usepackage{psfrag,subfigure,float}
\usepackage{multirow,subfigure,amsmath,epsfig,amsfonts,amssymb,psfig,graphics,psfrag,theorem,calc,subfigure,url,bm,cite}
\usepackage{stfloats,hyperref}
\usepackage{algorithm,algorithmic}
\usepackage{diagbox} 
\usepackage{hhline}
\usepackage{caption}
\usepackage{booktabs}
\usepackage{mathtools}

\def\blue{\color{blue}}
\def\red{\color{red}}

\usepackage[table,xcdraw]{xcolor}
\usepackage{colortbl} 

\makeatletter
\def\multilimits@{\bgroup
	\Let@
	\restore@math@cr
	\default@tag
	\baselineskip\fontdimen10 \scriptfont\tw@
	\advance\baselineskip\fontdimen12 \scriptfont\tw@
	\lineskip\thr@@\fontdimen8 \scriptfont\thr@@
	\lineskiplimit\lineskip
	\vbox\bgroup\ialign\bgroup\hfil$\m@th\scriptstyle{##}$\hfil\crcr}
\def\Sb{_\multilimits@}
\def\endSb{\crcr\egroup\egroup\egroup}
\makeatother

\definecolor{orange}{RGB}{255,107,0}
\definecolor{mediumseagreen}{rgb}{0.58, 0.44, 0.86}

\theorembodyfont{\rmfamily}

\newcommand\bA{\ensuremath{{\bm A}}}

\usepackage{graphicx}
\usepackage[export]{adjustbox}
\usepackage{bbding}
\usepackage{epsfig}
\usepackage{threeparttable}  
\usepackage{array}
\usepackage{makecell}
\usepackage[misc]{ifsym}
\usepackage[accsupp]{axessibility}

\definecolor{deepgreen}{rgb}{0.0, 0.6, 0.0} 
\ifCLASSINFOpdf
\else
\fi

\begin{document}

\title{Towards Robust DeepFake Detection under Unstable Face Sequences: Adaptive Sparse Graph Embedding with Order-Free Representation and Explicit Laplacian Spectral Prior}

\author{Chih-Chung~Hsu,~\IEEEmembership{Senior Member,~IEEE}, 
        Shao-Ning~Chen, 
        Mei-Hsuan~Wu,
        Chia-Ming~Lee, ~\IEEEmembership{Member,~IEEE},
        Yi-Fang~Wang, 
        and~Yi-Shiuan~Chou
        
\thanks{This study was supported in part by the National Science and Technology Council (NSTC), Taiwan, under grants MOST 112-2221-E-006-157-MY3 and 113-2627-M-006-005; and partly by the Higher Education Sprout Project of Ministry of Education (MOE) to the Headquarters of University Advancement at National Cheng Kung University (NCKU). We thank National Center for High-performance Computing (NCHC) of National Applied Research Laboratories (NARLabs) in Taiwan for providing computational and storage resources.}
\thanks{C.-C. Hsu is with Institute of Intelligent Systems, College of Artificial Intelligence and Green Energy, National Yang Ming Chiao Tung University, Hsinchu, Taiwan (e-mail: chihchung@nycu.edu.tw).}
\thanks{S.-N. Chen, M.-H. Wu, C.-M. Lee , Y.-F. Wang and Y.-S. Chou are with Institute of Data Science, National Cheng Kung University, Tainan, Taiwan (e-mail: johnnychen1999@gmail.com, zuw408421476@gmail.com, re6091054@gs.ncku.edu.tw, re6113018@gs.ncku.edu.tw, nelly910421@gmail.com).}
\thanks{Manuscript received XXX XX, 2025; revised XXX XX, 2025.}}

\markboth{IEEE Transactions on Information Forensics and Security, Vol.~XX, No.~X, MONTH~2025}%
{Hsu \MakeLowercase{\textit{et al.}}: Towards Robust DeepFake Detection}

\maketitle

\begin{abstract}
Ensuring the authenticity of video content remains challenging as DeepFake generation becomes increasingly realistic and robust against detection.
Most existing detectors implicitly assume temporally consistent and clean facial sequences, an assumption that rarely holds in real-world scenarios where compression artifacts, occlusions, and adversarial attacks destabilize face detection and often lead to invalid or misdetected faces.
To address these challenges, we propose a Laplacian-Regularized Graph Convolutional Network (LR-GCN) that robustly detects DeepFakes from noisy or unordered face sequences, while being trained only on clean facial data.
Our method constructs an Order-Free Temporal Graph Embedding (OF-TGE) that organizes frame-wise CNN features into an adaptive sparse graph based on semantic affinities. Unlike traditional methods constrained by strict temporal continuity, OF-TGE captures intrinsic feature consistency across frames, making it resilient to shuffled, missing, or heavily corrupted inputs.
We further impose a dual-level sparsity mechanism on both graph structure and node features to suppress the influence of invalid faces. Crucially, we introduce an explicit Graph Laplacian Spectral Prior that acts as a high-pass operator in the graph spectral domain, highlighting structural anomalies and forgery artifacts, which are then consolidated by a low-pass GCN aggregation. This sequential design effectively realizes a task-driven spectral band-pass mechanism that suppresses background information and random noise while preserving manipulation cues.
Extensive experiments on FF++, Celeb-DFv2, and DFDC demonstrate that LR-GCN achieves state-of-the-art performance and significantly improved robustness under severe global and local disruptions, including missing faces, occlusions, and adversarially perturbed face detections.

\end{abstract}


\section{Introduction}
\label{sec1}

Ensuring the authenticity of video content has emerged as a critical challenge amid rapid advances in DeepFake generation techniques. Early DeepFake videos often exhibited conspicuous artifacts and temporal discrepancies, enabling relatively straightforward detection. However, as generative adversarial networks (GANs), variational autoencoders (VAEs), and recent diffusion-based generative models have matured, malicious actors can now produce highly realistic manipulations that seamlessly integrate forged facial components, making their detection markedly more difficult. Such synthetic videos have far-reaching consequences, ranging from political misinformation and social engineering attacks to defamation and malicious entertainment. 

A critical yet often overlooked challenge in this context lies not only in the realism of the forgery itself, but in the instability of facial sequences in the wild. Video compression, adversarial attacks on face detectors, and natural occlusions frequently lead to missing, misaligned, or completely invalid facial crops, introducing severe noise and inconsistencies that can disrupt temporal modeling pipelines and significantly undermine the efficacy of existing detection methods.

A broad spectrum of solutions has been proposed to address the proliferation of DeepFakes. Image-based approaches often rely on deep CNN architectures \cite{chollet2017xception}, leveraging spatial patterns and subtle forgery traces within individual frames. Alternative strategies incorporate frequency-domain analyses \cite{qian2020thinking, fakefreq} or handcrafted features \cite{LRnet, fakecatcher} to capture anomalies that are not immediately visible in the RGB domain. Beyond spatial cues, temporal information has attracted growing attention. Methods such as \cite{LRnet}, \cite{fakefreq}, and \cite{RECCE} harness temporal inconsistencies to distinguish authentic facial movements from manipulated or stitched sequences. The Temporal Graph Convolutional Network (GCN) was introduced in \cite{yang2023masked} to model DeepFake detection as graph classification. However, existing temporal GCN methods typically assume a clear temporal ordering among frames, limiting their effectiveness when facing severe temporal disruptions or frame loss. Moreover, the availability of large-scale datasets, including FaceForensics++ (FF++) \cite{ffplus}, Celeb-DF \cite{celeb}, and the DeepFake Detection Challenge (DFDC) dataset \cite{dfdc}, has accelerated progress in both supervised and semi-supervised DeepFake detection frameworks \cite{hsu2020deep, hsuicip, hsu2}, while knowledge distillation \cite{kim2021fretal} and adversarial defense mechanisms \cite{cchsupami} have further enhanced the robustness and generalizability of certain approaches.

\begin{figure}
		\centering
		\includegraphics[width=0.4\textwidth]{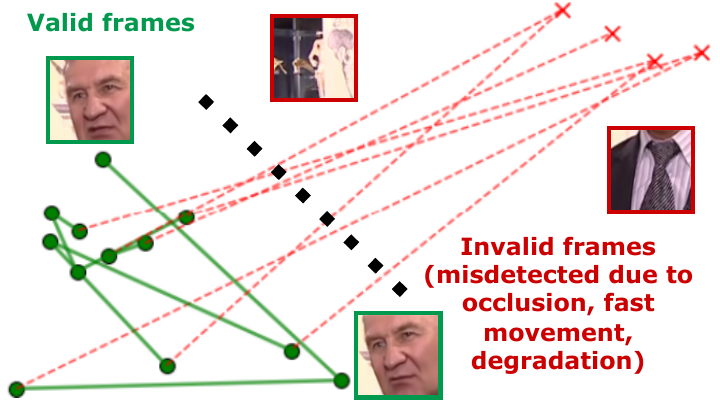}
		\caption{Illustration of realistic degradation scenarios in DeepFake detection, including invalid frames (red) caused by occlusions, rapid movements, or adversarial attacks, disrupting temporal feature trajectories (green).}\vspace*{-3mm}
		\label{fig:motivation}	\vspace*{-3mm}
\end{figure}

Despite these advancements, a critical assumption underpins most current DeepFake detection models: the input facial sequences are reliable and devoid of significant noise or inconsistencies. This assumption often falters in real-world scenarios where videos frequently exhibit various forms of degradation, as depicted in Figure~\ref{fig:motivation}. Such degradations can manifest globally, affecting the entire face, or locally, impacting specific facial regions. Global disruptions commonly result from video compression, blurring facial details or introducing artifacts, or from adversarial attacks~\cite{bose2018adversarial,attack1} designed to evade face detectors entirely. These disruptions produce invalid frames, which either fail face detection or mistakenly identify irrelevant objects as faces, significantly disrupting temporal feature consistency. Localized disruptions include occlusions (e.g., sunglasses or masks) or deliberate tampering with specific facial features, introducing spatial or temporal inconsistencies that severely hinder the detection process. These realistic perturbations highlight a crucial limitation of existing temporally dependent methods, underscoring the need for adaptive, order-free representations to robustly handle mixed or irregular degradation scenarios.

In addition to adversarial attacks and deliberate artifacts, facial sequences often exhibit natural instabilities arising from common real-world factors. Such instabilities frequently occur during video transmission, video compression, or from environmental occlusions. Unlike adversarial disruptions, which are intentionally crafted to fool detection models, these natural instabilities arise as unavoidable byproducts of daily video consumption and production. For instance, videos uploaded to social media platforms (e.g., TikTok, Instagram) undergo multiple rounds of compression, introducing blur and loss of detail. Similarly, face occlusion caused by masks, sunglasses, or hands can result in local feature inconsistencies. Video conferencing platforms (e.g., Zoom) also introduce frame drops or distortions due to unstable network conditions. Unlike controlled datasets, these natural disruptions are unavoidable in real-world deployments, underscoring the need for robust detection methods that generalize beyond ideal laboratory conditions. 

To address these inherent challenges, we propose an adaptive sparse graph embedding framework that explicitly abandons temporal ordering assumptions, a major limitation of existing temporal graph-based methods. Unlike prior graph-based DeepFake detectors that assume well-defined node relationships or stable embeddings \cite{yang2023masked}, we dispense with the notion of a strictly ordered temporal sequence and form an adaptive sparse graph whose nodes represent localized feature responses and edges encode their affinities based on feature similarities. This order-free, adaptive sparse representation liberates our model from the brittle dependencies of traditional temporal models, enabling robust detection even in sequences with missing or shuffled frames. Unlike traditional dynamic graph approaches that explicitly model temporal evolution, our adaptive sparse graph constructs edges solely based on adaptive feature affinities, adjusting connections without relying on temporal dependencies. Instead of collapsing when confronted with chaotic input, the network flexibly “rewires” its connections to preserve essential relationships, allowing us to handle large swaths of corrupted or missing data without losing track of discriminative facial patterns.

Yet, a graph structure alone doesn’t ensure resilience without dual sparsity on edges and features. Previous attempts that rely solely on graph connectivity risk inheriting the noise embedded in node features or in the graph’s topology. To tackle this, we impose a novel dual-level sparsity constraint. Unlike typical strategies that focus solely on making the graph itself sparse, we also enforce sparsity directly on the node features. By pruning non-informative signals at their source, we ensure that irrelevant or misleading traits introduced by invalid faces are discarded before they can propagate through the network. This dual sparsity—across both the graph and its node embeddings—goes beyond what earlier GCN- or TGCN-based solutions \cite{yang2023masked} attempt. Reducing the number of edges alone cannot fully neutralize the impact of severe noise if the remaining node features still carry destructive artifacts. Our design, on the other hand, guarantees that the graph’s backbone is not only structurally sparse but also filled with genuinely meaningful, high-quality features.

Even with an adaptively structured and feature-sparsified graph, distinguishing subtle forgery artifacts from random high-frequency noise (e.g., adversarial perturbations or sensor noise) remains challenging.
Traditional GCN formulations, which rely primarily on the adjacency matrix $\bA$ to aggregate information, inherently act as low-pass filters. While this smooths representations, it risks over-smoothing the subtle, high-frequency cues that are critical for identifying DeepFakes.
To address this, we introduce a \textbf{Laplacian-Regularized Graph Convolutional Network (LR-GCN)}, explicitly utilizing the graph Laplacian matrix.
Instead of merely smoothing, our Graph Laplacian Prior serves as a high-pass pre-filter, suppressing common low-frequency facial semantics (e.g., skin texture, illumination) and highlighting node-level inconsistencies.
Subsequently, the GCN aggregation acts as a low-pass filter to consolidate these highlighted cues.
This sequential combination effectively creates a learnable band-pass filter: it discards irrelevant background information (low frequency) and suppresses isolated random noise (ultra-high frequency), while selectively preserving the structural artifacts (mid-to-high frequency) inherent to manipulation.

Unlike conventional approaches that tackle distortions in face sequences by simulating specific corruptions (e.g., masked learning) during training, which often restrict model performance to seen distortion types and require extensive labeled data, our LR-GCN framework takes a different approach.
Instead of enlarging the supervision space with distortion-specific labels, we embed robustness directly into the representation: the adaptive sparse graph isolates severe outliers (e.g., misdetected frames), while the spectral band-pass mechanism distinguishes structured forgery cues from random interference.
Consequently, LR-GCN can be optimized in a standard supervised manner on uncorrupted data while still generalizing to a wide spectrum of unseen real-world distortions.

The main contributions of our work are as follows:
\begin{itemize}
    \item We propose a novel Laplacian-Regularized Graph Convolutional Network (LR-GCN) for DeepFake detection. By integrating an explicit Laplacian high-pass filter with GCN aggregation, we form a robust spectral band-pass mechanism that effectively isolates forgery traces from background semantics and random noise.
    \item We develop an adaptive sparse graph embedding that constructs connections based on feature affinities rather than temporal order, enabling robust detection even when face sequences are irregular or incomplete.
    \item We design a dual-level robustness mechanism with sparsity constraints and spectral filtering, which effectively suppresses noise propagation while preserving discriminative facial features.
    \item We demonstrate through extensive experiments that our approach significantly outperforms existing methods on benchmark datasets, particularly under challenging conditions where face detection may be unreliable.
\end{itemize}

The remainder of this paper is organized as follows. In Section~\ref{sec:relatedwork}, we discuss related works on DeepFake detection, focusing on methods that address compression, adversarial robustness, and temporal modeling. Section~\ref{sec:proposed_method} details the proposed framework, including the construction of the graph from spatiotemporal features, the integration of GCN with the Graph Laplacian Spectral Prior, and the implementation of the sparsity constraint. In Section~\ref{sec:experiments}, we present the experimental setup, comprehensive results, and in-depth analyses. Section~\ref{sec:conclusions} concludes our study and outlines potential directions for future work.

\section{Related Works}\label{sec:relatedwork}


\subsection{DeepFake Video Detection}\label{sec:generalrelatedwork}
In the domain of DeepFake video detection, numerous sophisticated approaches have recently emerged \cite{masi2020two,fakevideo1,fakexray,wang2022adt,LRnet,sabir2019recurrent,yang2019exposing,li2018ictu,fakecatcher,fakewarping}. First, these methods extend DeepFake image detection techniques by averaging the predictions of individual frames to assess a video's authenticity \cite{fakexray,li2018ictu,fakecatcher,fakewarping}. Second, the temporal inconsistency is exploited for DeepFake video detection using supervised learning approaches, as demonstrated in \cite{LRnet,sabir2019recurrent,masi2020two,fakevideo1}. Recently, several advanced techniques have been proposed to enhance DeepFake video detection performance. To address the generalizability issue, semi-supervised learning is considered in \cite{hsu2020deep, hsuicip, hsu2} to capture common fake features from selected representative GANs, assuming that most GANs might share similar identifiable clues, thereby improving generalizability for DeepFake image detection \cite{hsuicip, hsu2020deep}. 
CORE \cite{ni2022core} introduces a novel approach for learning consistent representations across different frames, while RECCE \cite{RECCE} employs a reconstruction-classification learning scheme to capture more discriminative features. DFIL \cite{pan2023dfil} proposes an incremental learning framework that exploits domain-invariant forgery clues to improve generalization ability. TALL-Swin \cite{xu2023tall} utilizes a thumbnail layout and Swin Transformer to learn robust spatiotemporal features for DeepFake detection. UCF \cite{yan2023ucf} focuses on uncovering common features shared by different manipulation techniques to enhance generalizability. DFGaze \cite{DFGAZE} uses gaze analysis of face video frames and then applies a spatiotemporal feature aggregator to realize authenticity classification.

\subsection{Graph Learning for DeepFake Detection}\label{sec:graphreatedwork}
Graph representation learning has emerged as a promising approach for DeepFake detection \cite{wang2023sfdg,she2024generalization,yang2023masked,khalid2023dfgnn}, offering robustness against structural degradation and irregularities. Unlike traditional CNNs, which focus on local patterns and neglect relational dependencies, and ViTs, which require extensive computational resources, GNNs excel at adaptively capturing high-order relationships among facial regions and across frames. 

Recent graph-based methods have leveraged relational structures within video sequences for DeepFake detection, but almost all of them implicitly assume clean and temporally ordered facial inputs. Wang et al. \cite{wang2023sfdg} proposed the Spatial-Frequency Domain Graph (SFDG), which constructs dynamic graphs guided by frequency-domain features and temporal ordering, improving detection performance under standard temporal sequences. She et al. \cite{she2024generalization} enhanced generalization by combining RGB features with auxiliary modalities and dynamically adjusting graph connectivity, yet still relying on temporally consistent node embeddings. Yang et al. \cite{yang2023masked} introduced Masked Relation Learning, employing edge-level sparsity to selectively preserve relations that explicitly depend on temporal ordering, and Khalid et al. \cite{khalid2023dfgnn} modeled multi-scale graph relationships. However, all these approaches inherently require stable frame sequences and reliable face detections, which makes them fragile when a large portion of frames are missing, corrupted, or misdetected.

While recent graph-based methods have advanced DeepFake detection, they share several fundamental limitations in terms of robustness. Most approaches implicitly rely on a clear temporal order and stable node representations and do not explicitly address adversarial attacks on face detectors or severe local corruptions, such as occlusions and regional noise. For example, SFDG \cite{wang2023sfdg} constructs dynamic graphs guided by spatial-frequency features but assumes reliable facial regions and standard temporal sequences. She et al. \cite{she2024generalization} improve generalization by representing each image as a graph and dynamically adjusting connectivity, yet they still depend on clean, complete facial regions. Masked Relation Learning \cite{yang2023masked} imposes edge-level sparsity to selectively preserve relations that explicitly depend on temporal ordering, and DFGNN \cite{khalid2023dfgnn} models multi-scale graph relationships; however, both inherently require stable frame sequences and reliable face detections.

When faced with global disruptions, such as severe degradation or occlusions that prevent reliable face detection, these methods often degrade substantially, as they rely heavily on pre-detected facial regions and ordered temporal trajectories to construct their graphs. Such vulnerabilities highlight their fragility in handling real-world scenarios with missing data or heavily perturbed sequences. In contrast, our adaptive sparse graph embedding is explicitly designed to operate without temporal ordering constraints and to remain effective even when many frames are invalid or corrupted.

To overcome these limitations, we propose an adaptive sparse graph embedding framework explicitly designed without temporal ordering constraints, a critical departure from existing approaches \cite{gcndfd, wang2023sfdg, yang2023masked, khalid2023dfgnn}. Our graph representation adaptively constructs relationships based on learned CNN feature affinities, thus robustly accommodating severely corrupted, unordered, or partially missing facial sequences. Additionally, we introduce a novel dual-level sparsity constraint to simultaneously prune irrelevant graph connections and redundant node features. Finally, we propose an explicit GLSP integrated into our LR-GCN, which shapes the graph representation via a Laplacian-based high-pass pre-filter followed by GCN-based low-pass aggregation, effectively stabilizing node embeddings by suppressing irrelevant background semantics and random noise while preserving discriminative forgery artifacts. This cohesive framework enables our method to substantially outperform existing state-of-the-art graph-based approaches under challenging real-world conditions.

\section{Laplacian-Regularized Graph Convolutional Network}\label{sec:proposed_method}

\begin{figure*}
		\centering
		\includegraphics[width=0.95\textwidth]{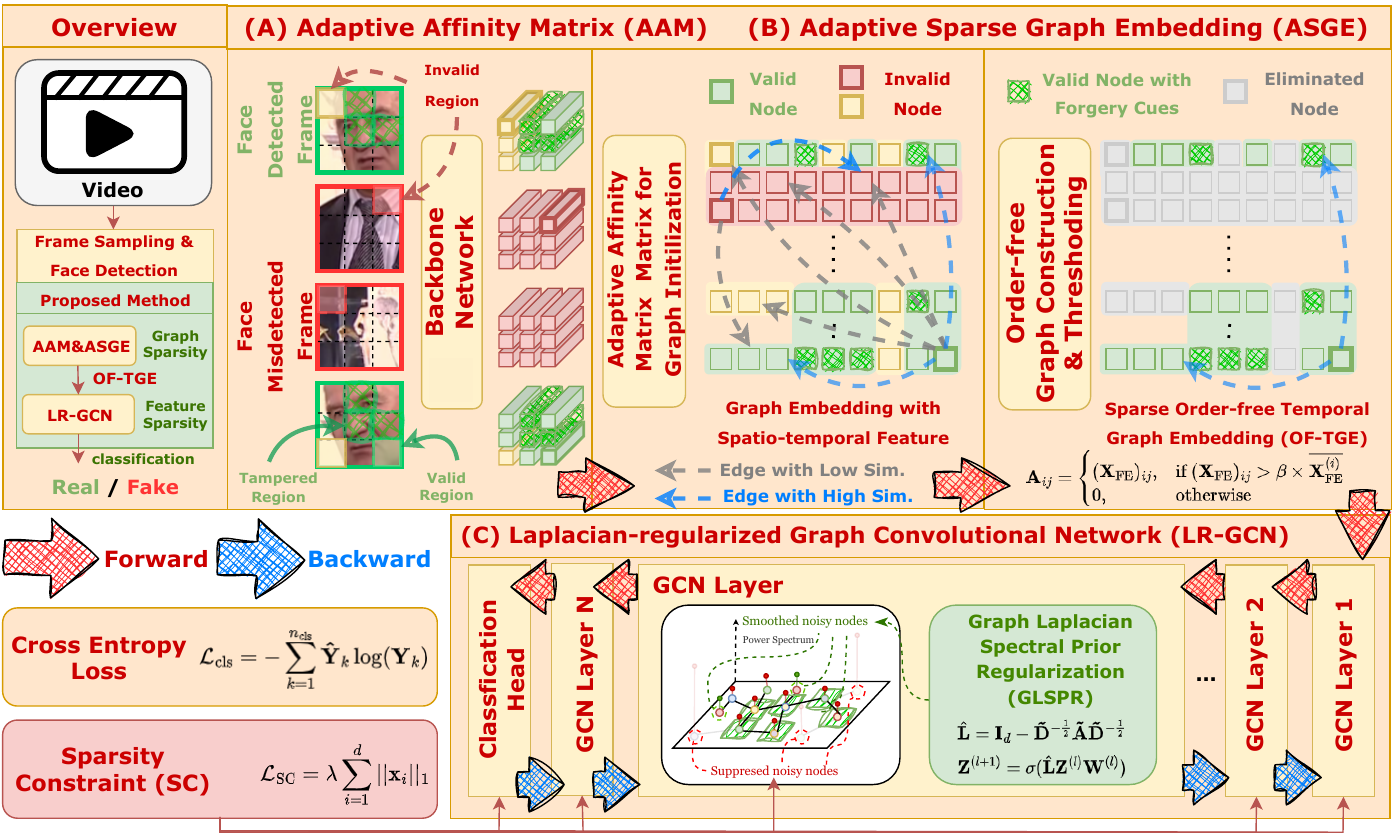}
		\caption{Flowchart of the proposed LR-GCN framework for robust DeepFake video detection. Due to unreliable face detection, invalid faces often significantly outnumber valid ones, causing traditional DeepFake detection methods to degrade. Our method employs an Adaptive Sparse Graph Embedding (ASGE) to structurally isolate severe outliers (e.g., misdetected or background frames) and then applies a spectral band-pass mechanism that combines an explicit Laplacian high-pass pre-filter (to highlight forgery artifacts and structural inconsistencies) with GCN-based low-pass aggregation (to consolidate consistent evidence and suppress isolated noise), all under dual-level sparsity constraints on both graph structure and node features. This three-stage spectral sieving design enables LR-GCN to robustly handle noisy and corrupted facial sequences.}\vspace*{-2mm}
		\label{fig:features}	\vspace*{-2mm}
\end{figure*}

\subsection{Overview of the Proposed Method}\label{sec:overview}
A key distinction of our LR-GCN framework lies in its independence from simulating specific distortions during training, a common practice in traditional supervised methods. Instead, by constructing an adaptive sparse graph and incorporating explicit Laplacian spectral priors, our approach naturally reduces the influence of unstable face sequences. This design eliminates the need for distortion-specific labeled data, offering a robust solution that generalizes effectively to a wide range of real-world corruptions.

To address challenges posed by unstable face sequences with global distortions (e.g., jittering, frame drops) and local corruptions (e.g., occlusions, adversarial attacks), we propose a robust Adaptive Sparse Graph Embedding (ASGE) framework. Specifically, our framework explicitly discards fixed temporal ordering constraints by leveraging an Order-Free Temporal Graph Embedding (OF-TGE). As illustrated in Figure~\ref{fig:features}, we first perform face detection and CNN-based feature extraction (e.g., ResNet, EfficientNet) to obtain spatially localized facial features from each detected face region. Recognizing that real-world sequences often contain corrupted or invalid frames, OF-TGE organizes these spatial features into an order-free graph representation.

The extracted CNN features are organized into a unified feature matrix $\mathbf{X} $, where each node represents localized facial semantics. We compute an adaptive affinity matrix $\mathbf{A}$ based on semantic similarities measured by inner products between CNN features ($\mathbf{X}\mathbf{X}^T$). An adaptive thresholding strategy selectively sparsifies these graph connections, retaining only the most informative edges and discarding weak or irrelevant links.

To further enhance robustness, we introduce a dual-level sparsity constraint (SC) at the node-feature level. By applying $\ell_1$ regularization directly to node features, we encourage feature-level sparsity and systematically prune less informative or corrupted node signals, preventing invalid faces from dominating the representation. Even with dual-level sparsity constraints, however, simply relying on standard GCN aggregation is insufficient to discriminate subtle forgery artifacts from random high-frequency perturbations. To tackle this, we explicitly incorporate a Graph Laplacian Spectral Prior (GLSP) into our GCN, termed LR-GCN (Laplacian-Regularized GCN), leveraging the normalized graph Laplacian $\hat{\mathbf{L}} = \mathbf{I}_d - \tilde{\mathbf{D}}^{-\frac{1}{2}} \tilde{\mathbf{A}} \tilde{\mathbf{D}}^{-\frac{1}{2}}$, where $\tilde{\mathbf{A}} = \mathbf{A} + \mathbf{I}_d$ and $\tilde{\mathbf{D}}$ is its degree matrix. The Laplacian operator first acts as a fixed high-pass filter that suppresses common low-frequency background semantics (e.g., facial identity and global lighting) and highlights inconsistencies between semantically related nodes, while the subsequent GCN propagation behaves as a learnable low-pass filter that consolidates consistent mid-frequency responses and attenuates isolated spikes. Together with the
With a feature-level sparsity constraint, this sequential design implements a task-driven spectral band-pass mechanism that significantly stabilizes node embeddings under severe perturbations.

Integrating ASGE, dual-level sparsity constraints, and explicit GLSP into our LR-GCN, we achieve superior robustness, significantly outperforming existing methods under challenging real-world conditions.

\subsection{Adaptive Sparse Graph Embedding}\label{sec:of_tge}

Most current DeepFake detection methods struggle with unstable face sequences, whereas our ASGE constructs the graph in an order-free manner, leading to better performance. Let us denote the spatial feature map extracted from a single frame as $\mathbf{F} \in \mathbb{R}^{C \times h \times w}$, where $C$ represents the number of channels, and $h$ and $w$ denote the height and width of the spatial dimensions of the feature map, respectively. The feature vector at a specific spatial location $(i, j)$ within frame $n$ can be represented as $\mathbf{f}_{n, i, j} \in \mathbb{R}^{C \times 1}$. For a video with $N$ frames, a naive approach would be to concatenate these spatial features across all spatial locations and frames to form a large spatiotemporal feature matrix. However, this approach suffers from several drawbacks. Firstly, it imposes a strict temporal ordering, making it vulnerable to frame disorder or missing frames, which are common in real-world videos. Secondly, processing this high-dimensional feature matrix directly, especially with methods like Transformers \cite{videoTrans}, leads to significant computational complexity due to the large number of tokens and the need to compute long-range dependencies.

To address these limitations, we propose the OF-TGE, a novel approach that constructs an adaptive graph representation of the video without relying on a fixed temporal order. Instead of concatenating features directly, we treat each spatial feature vector $\mathbf{f}_{n, i, j}$ as a node in our graph. This allows us to capture relationships between different spatial locations and time instances in an order-free manner. We organize these node features into a feature matrix $\mathbf{X} \in \mathbb{R}^{d \times C}$, where $d = N \times h \times w$ is the total number of nodes, and each row $\mathbf{x}_k \in \mathbb{R}^{1 \times C}$ of $\mathbf{X}$ corresponds to a feature vector $\mathbf{f}_{n, i, j}$, with $k$ indexing across all spatial locations and frames.

To capture the relationships between these nodes, we simply compute an Adaptive Affinity Matrix (AAM) $\mathbf{X}_{\text{AAM}} \in \mathbb{R}^{d \times d}$ as follows:

\begin{equation}
    \mathbf{X}_{\text{AAM}} = \mathbf{X} \mathbf{X}^T.
\end{equation}

The element $(\mathbf{X}_{\text{AAM}})_{pq}$ represents the inner product between the feature vectors of nodes $p$ and $q$, quantifying their semantic similarity, as their edge weights. This inner product effectively captures the correlation between different spatial locations across different frames, without being constrained by temporal order. The underlying hypothesis is that authentic videos exhibit strong semantic consistency across frames (high affinity) regardless of order, whereas manipulated videos introduce structural inconsistencies that disrupt these affinities, particularly in the feature space constructed by the backbone. For instance, the element $(\mathbf{X}_{\text{AAM}})_{1, d}$ captures the relationship between the first spatial location in the first frame and the last spatial location in the last frame, directly capturing potential long-range dependencies.

To focus on the most salient relationships and reduce computational complexity, we apply an adaptive thresholding strategy to $\mathbf{X}_{\text{AAM}}$ to create our ASGE. For $i$-th node, we calculate the average affinity
$\overline{\mathbf{X}_{\text{AAM}}^{(i)}} = \frac{1}{d}\sum_{j=1}^{d} (\mathbf{X}_{\text{AAM}})_{ij}$. The elements of the affinity matrix $\mathbf{A}$ are then determined as follows:

\begin{equation}\label{eq:a}
    \mathbf{A}_{ij} =
    \begin{cases}
        (\mathbf{X}_{\text{AAM}})_{ij}, & \text{if } (\mathbf{X}_{\text{AAM}})_{ij} > \beta \times \overline{\mathbf{X}_{\text{AAM}}^{(i)}} \\
        0, & \text{otherwise},
    \end{cases}
\end{equation}
where $\beta$ is a hyperparameter controlling the sparsity level and we set $\beta=0.5$ for all experiments (see Appendix to find the hyperparameter $\beta$ selection). This adaptive thresholding, based on the local context of each node, allows us to adaptively filter out weak or spurious connections, leading to a more robust and efficient graph representation. After hresholding, we further symmetrize the adjacency as $\mathbf{A} \leftarrow (\mathbf{A} + \mathbf{A}^\top)/2$ to obtain an undirected graph, which ensures that the normalized Laplacian in Section~III-D is real and symmetric and thus admits a
standard spectral interpretation.

This ASGE, based on the proposed Order-Free Temporal Graph Embedding (OF-TGE), provides a crucial foundation for subsequent processing by allowing us to work with a graph structure that is invariant to frame order and robust to noisy nodes, while efficiently capturing relevant spatiotemporal relationships. In particular, by discarding explicit temporal indices and relying solely on feature affinities, OF-TGE naturally adapts to videos where valid frames are sparse, shuffled, or heavily interleaved with invalid detections.

\subsection{Sparsity Constraint of Node Features}\label{sec:sc}

As discussed in Section III-B, ASGE plays a crucial role in establishing a robust graph structure by adaptively filtering weak or spurious connections. However, ASGE primarily focuses on relationships between nodes, operating at the graph-structure level. Noisy features—especially those arising from structured or correlated noise—may still exhibit strong affinities with other nodes and thus survive the structural pruning. This motivates a complementary mechanism that operates directly on node features.

To this end, we introduce a Sparsity Constraint (SC) on the feature matrix $\mathbf{X} \in \mathbb{R}^{d \times C}$, where each row $\mathbf{x}_i \in \mathbb{R}^{1 \times C}$ represents the feature vector of node $i$. The goal is to suppress the contribution of less informative or corrupted responses, regardless of their initial energy or connectivity within the graph, as illustrated in Fig. \ref{fig:oftge}. By driving many feature dimensions towards zero, SC encourages the model to rely on a compact set of salient, noise-resilient cues.

We implement SC via an $\ell_1$ regularization term applied to node features. The $\ell_1$ norm of a vector $\mathbf{x}$, defined as $\lVert \mathbf{x} \rVert_1 = \sum_{j=1}^{C} |x_j|$, is well known to promote sparsity by shrinking a large portion of the coordinates exactly to zero. This property is particularly suitable for mitigating noise distributed across multiple feature dimensions. Formally, the sparsity loss is defined as
\begin{equation}
\mathcal{L}_{\text{SC}} = \lambda \sum_{i=1}^{d} \left\|
\mathbf{x}_i \right\|_1 ,
\label{eq:sc_loss}
\end{equation}
where $\lambda$ is a hyperparameter controlling the strength of the sparsity constraint.

During training, the SC term is combined with the standard cross-entropy classification loss:
\begin{equation}
  \mathcal{L}
  = -\sum_{k=1}^{n_{\text{cls}}} \mathbf{Y}_{k} \log \hat{\mathbf{Y}}_{k}
    + \mathcal{L}_{\text{SC}},
  \label{eq:total_loss}
\end{equation}
where $\mathbf{Y}$ denotes the one-hot ground-truth label and
$\hat{\mathbf{Y}}$ is the predicted probability distribution over
$n_{\text{cls}}$ classes for each sample, and the summation is taken over the class dimension. In practice, $\mathcal{L}_{\text{SC}}$ is computed from the concatenated spatio-temporal feature matrix $\mathbf{X}$ after OF-TGE, so that all nodes across frames and spatial locations jointly contribute to the sparsity penalty. Nodes associated with invalid or weakly informative regions tend to receive smaller gradients from the classification loss and are therefore more heavily shrunk by the $\ell_1$ term, gradually driving their feature responses towards zero. This implementation is equivalent to applying an $\ell_1$ penalty on all node features and yields a compact representation in which only a small subset of nodes carries non-negligible activations. Compared to pruning edges alone, this feature-level sparsity explicitly suppresses activations of invalid or weakly informative nodes before message passing, providing a robust and complementary mechanism to mitigate the impact of noise in DeepFake detection.

\begin{figure}
		\centering
		\includegraphics[width=0.47\textwidth]{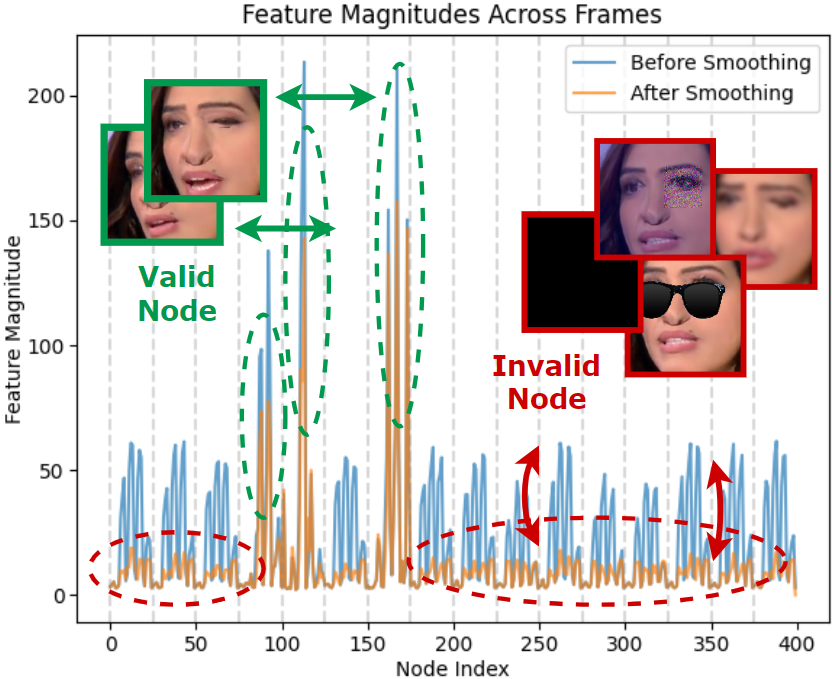}
            \caption{Visualization of node-wise feature magnitudes across frames. Valid nodes (green) exhibit consistently strong activations, while invalid nodes (red) corrupted by occlusion, blur,
or adversarial noise—produce scattered, low-magnitude responses. The proposed dual-level sparsity, together with the Graph Laplacian Spectral Prior and subsequent GCN-based aggregation, effectively realizes a spectral band-pass behavior that suppresses noisy activations from invalid nodes while preserving and stabilizing the discriminative responses of valid nodes, enabling a robust, order-free representation.}
            		\vspace*{-2mm}\label{fig:oftge}	\vspace*{-2mm}
\end{figure}

\subsection{Explicit Graph Laplacian Spectral Prior}

Although the preceding components of our framework—namely the adaptive sparse graph embedding (Section~III-B), edge-level sparsity, and feature-level sparsity (Section~III-C)—remove many noisy nodes and suppress uninformative features, distinguishing between random perturbations and genuine forgery artifacts requires spectral discrimination.
Standard GCNs implicitly act as low-pass filters, smoothing node features by aggregating information from neighbors. While this suppresses noise, it may also wash out subtle high-frequency forgery clues.
To resolve this, we embed a graph Laplacian-based prior into the GCN in the spirit of graph signal processing (GSP)~\cite{gsp,gsptutorial}.

We briefly recall the standard GCN propagation rule~\cite{gcn1,gcn2}. Given a graph with adjacency matrix $\mathbf{A}$ and node features $\mathbf{Z}^{(l)}$ at layer $l$, the next-layer features are computed as
\begin{equation}
    \mathbf{Z}^{(l+1)} = \sigma\!\left(
    \tilde{\mathbf{D}}^{-\frac{1}{2}}
    \tilde{\mathbf{A}}
    \tilde{\mathbf{D}}^{-\frac{1}{2}}
    \mathbf{Z}^{(l)} \mathbf{W}^{(l)}
    \right),
    \label{eq:gcn_standard}
\end{equation}
where $\tilde{\mathbf{A}} = \mathbf{A} + \mathbf{I}_d$ denotes the adjacency matrix with self-loops and $\tilde{\mathbf{D}}$ is the corresponding degree matrix.
This operation essentially performs low-pass filtering, making node representations similar to their neighbors.

To extract discriminative high-frequency cues before aggregation, we construct the normalized graph Laplacian:
\begin{equation}
  \hat{\mathbf{L}} = \mathbf{I}_d - \tilde{\mathbf{D}}^{-\frac{1}{2}}
    \tilde{\mathbf{A}} \tilde{\mathbf{D}}^{-\frac{1}{2}}.
  \label{eq:laplacian}
\end{equation}
In GSP, $\hat{\mathbf{L}}$ acts as a difference operator. Its eigenvalues $\lambda_k \in [0,2]$ index frequency components: small $\lambda_k$ correspond to smooth (low-frequency) variations, while large $\lambda_k$ reflect rapid (high-frequency) fluctuations across connected nodes.
Our method introduces an explicit Laplacian pre-filtering
stage:
\begin{equation}
    \mathbf{Z}^{(0)}_{\text{lap}} = 
    \hat{\mathbf{L}} \mathbf{X},
    \label{eq:lap_pre}
\end{equation}
where $\mathbf{X}$ is the node-feature matrix produced by ASGE and $\mathbf{Z}^{(0)}_{\text{lap}}$ serves as the input to the first GCN layer, i.e., we set $\mathbf{Z}^{(0)} = \mathbf{Z}^{(0)}_{\text{lap}}$ in \eqref{eq:gcn_standard}. 

Unlike smoothing, Eq.~\eqref{eq:lap_pre} acts as a \textit{high-pass filter}. Since $\hat{\mathbf{L}}\mathbf{x} \approx \mathbf{x} - \text{smoothed}(\mathbf{x})$, this operation effectively suppresses the common low-frequency background information (e.g., facial identity and global lighting) and highlights the \textit{inconsistencies} between semantically related nodes.
In the context of DeepFake detection, these high-frequency residues are critical as they often harbor subtle forgery artifacts or structural discontinuities caused by manipulation.

However, high-frequency signals can also contain random noise. This is where the subsequent GCN layers play a complementary role: the GCN propagation in Eq.~\eqref{eq:gcn_standard} aggregates these highlighted high-frequency cues.
The GCN propagation in Eq.~\eqref{eq:gcn_standard} aggregates these highlighted high-frequency cues.
If a high response in $\mathbf{Z}^{(0)}_{\text{lap}}$ is isolated (random noise), the GCN aggregation tends to suppress it because of lack of neighbor support.
Conversely, if the high-frequency response is structurally consistent among neighbors (e.g., a manipulated region boundary or occlusion edge), the GCN consolidates and reinforces this evidence.
Consequently, the cascading of the Laplacian pre-filter (high-pass) and the subsequent GCN propagation (low-pass) effectively constitutes a task-driven spectral band-pass mechanism, which is further supported by the ablation results in Table~\ref{tab:module}, where enabling GLSP and SC yields consistent gains under high masking ratios.
It filters out irrelevant global background (very low freq) and suppresses random jitter (ultra-high freq), while selectively focusing on the structural artifacts (mid-to-high freq) that are most discriminative for DeepFake detection.

\subsection{Classification Learning}

Finally, the output features are obtained by passing the last GCN layer's embeddings $\mathbf{Z}^{(L)}$ through a fully connected (FC) layer:
\begin{equation}
\mathbf{Z} = \sigma\big( \mathbf{Z}^{(L)} \mathbf{W}_{\text{out}} \big) ,
\end{equation}
where $\mathbf{W}_{\text{out}} \in \mathbb{R}^{g_{\text{dim}} \times n_{\text{out}}}$ denotes the weight matrix of the FC layer, $g_{\text{dim}}$ is the embedding dimension of the GCN, and $n_{\text{out}}$ is the number of neurons in the FC layer. The predicted class probabilities are then given by
\begin{equation}
\hat{\mathbf{Y}} = \text{Softmax}\big( \mathbf{Z}\, \mathbf{W}_{\text{cls}} \big) ,
\end{equation}
where $\mathbf{W}_{\text{cls}} \in \mathbb{R}^{n_{\text{out}} \times n_{\text{cls}}}$ denotes the classification weight matrix and $n_{\text{cls}}$ is the number of classes. The overall loss in~(\ref{eq:total_loss}) is used to train the entire LR-GCN model in an end-to-end manner.

\section{Experimental Results}\label{sec:experiments}
\subsection{Experimental Configuration}
\textbf{Dataset setup.} The robustness validation of the proposed method is the core of our investigation, particularly when applied to noisy face sequences containing many invalid faces. To achieve this, representative benchmark datasets are essential. Therefore, we selected three well-established benchmark datasets for performance evaluation: FF++ \cite{ffplus}, Celeb-DFv2 \cite{celeb}, and the large-scale DFDC dataset \cite{dfdc}. The FF++ dataset \cite{ffplus} comprises four distinct classes of manipulation methods: 1) DeepFakes (DF), 2) Face2Face (F2F), 3) FaceSwap (FS), and 4) NeuralTextures (NT). For each class, a set of 1,000 original videos was used to generate 1,000 manipulated versions, resulting in a total of 1,000 authentic and 4,000 doctored videos. The Celeb-DF dataset \cite{celeb} contains 590 original videos and 5,639 manipulated counterparts, generated using improved generative adversarial networks at a resolution of $256\times 256$. To enhance the quality of manipulated videos, Celeb-DF \cite{celeb} employs a Kalman filter to mitigate temporal inconsistencies across successive frames. The DFDC dataset \cite{dfdc}, created by Facebook in collaboration with other organizations, is a large-scale dataset designed to facilitate the development of DeepFake detection algorithms. It consists of over 100,000 videos, featuring a mix of authentic and manipulated content generated using various state-of-the-art face-swapping and facial reenactment techniques, ensuring a diverse and challenging set of DeepFakes for evaluation.

\textbf{Training Hyperparameters of LR-GCN.}
To ensure balanced performance appraisal, the FF++ \cite{ffplus}, Celeb-DF \cite{celeb}, and DFDC \cite{dfdc} datasets were split into training, validation, and test sets at a $8:1:1$ ratio. In line with our objective to ascertain the efficacy of LR-GCN in the presence of unstable face detectors, we trained separate LR-GCN models independently on each dataset. During the training phase, the Adam optimizer \cite{kingma2017adam} was used with an initial learning rate of $1e^{-4}$ and a stepwise learning rate decay schedule. We employed the 53-layer Cross Stage Partial Network (CSPNet) \cite{bochkovskiy2020yolov4} as the backbone network. Note that any CNNs could be used in our LR-GCN as backbone network. The standard GCN with our Graph Laplacian was implemented for stacking $g_n$ layers, with $g_n=8$ and an embedding size of $g_\text{dim}=400$ as the default settings in this study. The number of neurons of the last fully connected layer $n_\text{out}$ is $2,048$ for our experiments. All facial images were resized to $144 \times 144$ during both training and inference stages. Standard data augmentation techniques, such as random noise, cropping, and flipping, were applied during training. The training phase consisted of 200 epochs, with a learning rate decay of 0.1 every 100 epochs. We randomly sampled $N=16$ successive facial images to form the input tensor for our experiments. All comparison methods, including the proposed LR-GCN, were trained on the training set and evaluated on the testing set.

\textbf{Training Hyperparameters of Peer Methods.}
For performance evaluation, we compared our proposed method with several state-of-the-art DeepFake detection techniques, including Xception \cite{chollet2017xception}, $F^3$-net \cite{qian2020thinking}, RECCE \cite{RECCE}, DFIL \cite{pan2023dfil}, UCF \cite{yan2023ucf}, CORE \cite{ni2022core}, TALL-Swin \cite{xu2023tall}, DFGAZE  \cite{DFGAZE} and MaskRelation \cite{yang2023masked}. The image-based approaches, namely Xception, $F^3$-net, RECCE, DFIL, UCF, and CORE, were trained using the same strategy as described previously, with their default settings. However, the learning rates of Xception \cite{chollet2017xception} and UCF \cite{yan2023ucf} were adjusted to $2e^{-4}$ for better performance. The video-based approaches TALL-Swin \cite{xu2023tall} and DFGAZE \cite{DFGAZE} were trained with their default settings. During the training phase, we randomly selected $N$ facial images from the training set. The final authenticity verdict for the input video was determined by averaging the $N$ prediction outcomes corresponding to the $N$ facial images extracted from the input video, using a temporally centered cropping strategy. For all other methods, the number of frames $N$ used was set to $16$. The image size for Xception \cite{chollet2017xception}, $F^3$-net \cite{qian2020thinking}, RECCE \cite{RECCE}, UCF \cite{yan2023ucf}, and CORE \cite{ni2022core} is $256\times 256$, suggested by their default settings, while that for DFIL \cite{pan2023dfil}, TALL-Swin \cite{xu2023tall}, and DFGAZE \cite{DFGAZE} are $299\times 299$, $224\times 224$, and $224\times 224$, respectively. 

\textbf{Settings in Inference Phase.}
To evaluate the model's performance under the influence of an unstable face detector, we randomly replaced certain facial images with background segments, as determined by the masking ratio $m_r$. We experimented with masking ratios ranging from 0.1 to 0.8 to assess the effectiveness of LR-GCN under varying levels of noise in the face sequences. For instance, with $N=16$ and $m_r=0.5$, up to eight facial images could be replaced with background images in the corresponding frames, simulating real-world scenarios where face detection may be challenging or unreliable.
In our experimental setup, we sampled $N=16$ frames from the middle portion of each video, using the same approach as during training. When $m_r=0.5$, half of the 16 frames (i.e., 8) were randomly replaced with either background or completely black images. By varying the masking ratio, we evaluated the robustness and stability of each method under different levels of noise in the face sequences.

Furthermore, we assumed that each frame should contain at least one face to simulate adversarial attacks on face detectors in real-world scenarios. In cases where no face was detected in a frame, we replaced that frame with a black image, generating a noisy face sequence that allowed us to assess the robustness of LR-GCN under challenging conditions.
Our experimental analysis employed three performance metrics: accuracy, macro F1-Score, and Area Under the Receiver Operating Characteristic Curve (AUC). 
For simplicity, these metrics are referred to as Accuracy (Acc.), F1-Score, and AUC throughout the experimental sections.

\subsection{Quantitative Results}
\begin{table*}[]
    \centering\small
    \caption{Quantitative comparison of the noisy face sequences under different masking ratios $m_r$ between the proposed LR-GCN and other state-of-the-art methods. We highlight the best performance in red and the second-best performance in blue, considering the several benchmark dataset, such as FF++ \cite{ffplus}, Celeb-DF \cite{celeb}, and DFDC \cite{dfdc} with different $m_r$. 
}
    \scalebox{0.8}{
    \begin{tabular}{c|c|c|c|ccc|ccc|ccc|ccc}
    \hline
        \hline
        \multirow{2}{*}{Method} & \multirow{2}{*}{Venue} & \multirow{2}{*}{Type} &  \multirow{2}{*}{$m_r$} & \multicolumn{3}{c}{FF++ \cite{ffplus}} & \multicolumn{3}{c}{Celeb-DF \cite{celeb}} & \multicolumn{3}{c}{DFDC \cite{dfdc}} & \multicolumn{3}{|c}{Average}\\ \cline{5-16} &  &  &  & Acc. & F1 & AUC & Acc. & F1 & AUC & Acc. & F1 & AUC & Acc. & F1 & AUC \\
        \hline
        \multirow{3}{*}{Xception \cite{chollet2017xception}} & \multirow{3}{*}{CVPR2017} & \multirow{3}{*}{Image-based} & 
        0.0 & 0.925 & 0.894 & 0.972 & 0.861 & 0.806 & 0.910 & 0.953 & 0.910 & 0.981 & 0.913& 0.870& 0.954
\\ & & & 
        0.4 & 0.869 & 0.780 & 0.871 & 0.631 & 0.614 & 0.782 & {\blue 0.908} & {\blue 0.788} & 0.866 & 0.803& 0.727& 0.840
\\ & & & 
        0.8 & 0.814 & 0.594 & 0.654 & 0.398 & 0.389 & 0.604 & {\blue 0.864} & 0.598 & 0.647 & 0.692& 0.527& 0.635\\
        \hline 
        \multirow{3}{*}{\text{$F^3$}-net \cite{qian2020thinking}} & \multirow{3}{*}{ECCV2020} & \multirow{3}{*}{Image-based} & 
        0.0 & 0.950 & 0.928 & 0.986 & {\blue 0.965} & {\blue 0.957} & {\blue 0.993} & {\blue 0.957} & {\blue 0.921} & {\blue 0.986} & {\blue 0.958} & {\blue 0.935} & {\blue 0.988} 
\\ & & & 
        0.4 & {\blue 0.883} & 0.798 & 0.888 & 0.691 & 0.684 &  0.895 &  0.864 & 0.655 & 0.755 & 0.813& 0.712& 0.846
\\ & & & 
        0.8 & 0.818 & 0.599 & 0.662 & 0.418 & 0.407 & 0.664 & 0.850 & 0.539 &  0.595 & 0.696& 0.515& 0.640\\
        \hline                                                       
        \multirow{3}{*}{RECCE \cite{RECCE}} & \multirow{3}{*}{CVPR2022} & \multirow{3}{*}{Image-based} & 
        0.0 & 0.938 & 0.911 & 0.979 & 0.941 & 0.925 & 0.985 & 0.940 & 0.872 & 0.973 & 0.940& 0.903& 0.979
\\ & & & 
        0.4 & 0.878 & 0.790 & 0.874 & 0.678 & 0.669 & 0.869 & 0.900 & 0.752 & 0.863 & 0.819& 0.737& 0.869
\\ & & & 
        0.8 & 0.817 & 0.599 & 0.655 & 0.414 & 0.404 & 0.648 & 0.861 & 0.579 & 0.648 & 0.698& 0.527& 0.650\\
        \hline
        \multirow{3}{*}{CORE \cite{ni2022core}} & \multirow{3}{*}{CVPRW2022} & \multirow{3}{*}{Image-based} &
        0.0 & 0.948 & 0.925 & 0.984 & 0.953 & 0.940 & 0.989 & 0.950 & 0.903 & 0.977 & 0.950& 0.922& 0.983
\\ & & & 
        0.4 & 0.883 & 0.799 & 0.888 & {\blue 0.858} & {\blue 0.790} & {0.890} & {0.907} & 0.781 & 0.870 & {\blue 0.882} & {\blue 0.790} & 0.883
\\ & & & 
        0.8 & 0.818 & 0.601 & 0.663 & 0.764 & 0.572 & 0.661 & 0.863 & 0.595 & 0.651 & {\blue 0.815} & 0.589& 0.658\\
        \hline
        \multirow{3}{*}{UCF \cite{yan2023ucf}} & \multirow{3}{*}{CVPR2023} & \multirow{3}{*}{Image-based} & 
        0.0 & 0.937 & 0.911 & 0.982 & 0.856 & 0.792 & 0.891 & 0.890 & 0.815 & 0.939 & 0.894& 0.840& 0.937
\\ & & & 
        0.4 & 0.875 & 0.790 & 0.882 & 0.626 & 0.607 & 0.642 & 0.871 & 0.733 & 0.812 & 0.791& 0.710& 0.779
\\ & & & 
        0.8 & 0.815 & 0.598 & 0.660 & 0.397 & 0.389 & 0.516 & 0.851 & 0.586 & 0.620 & 0.688& 0.524& 0.599\\
        \hline
        \multirow{3}{*}{DFIL \cite{pan2023dfil}} & \multirow{3}{*}{ACMMM2023} & \multirow{3}{*}{Image-based} & 
        0.0 & {\blue 0.954} & {\blue 0.939} & {\blue 0.987} & 0.957 & 0.954 & 0.964 & 0.940 & 0.881 & 0.955 & 0.950& 0.925& 0.969
\\ & & & 
        0.4 & 0.876 & 0.808 & 0.893 & 0.695 & 0.684 & 0.825 & 0.886 & 0.720 & 0.813 & 0.819& 0.737& 0.844
\\ & & & 
        0.8 & 0.759 & 0.603 & 0.665 & 0.518 & 0.350 & 0.644 & 0.855 & 0.565 & 0.621 & 0.711& 0.506&  0.644\\
        \hline
        \multirow{3}{*}{TALL-Swin \cite{xu2023tall}} & \multirow{3}{*}{ICCV2023} & \multirow{3}{*}{Video-based} & 
        0.0 & 0.913 & 0.868 & 0.881 & 0.913 & 0.933 & 0.924 & 0.911 & 0.812 & {0.984} & 0.912& 0.871& 0.930
\\ & & & 
        0.4 & 0.867 & 0.767 & 0.740 & 0.847 & 0.789 & 0.825 & 0.872 & 0.758 & {0.786} & 0.862& 0.771& 0.784
\\ & & & 
        0.8 & {\blue 0.827} & 0.605 &  0.589 & 0.745 & {\blue 0.680} & 0.645 & 0.845 & {\blue 0.688} & 0.650 & 0.806& 0.658& 0.628\\
        \hline
        \multirow{3}{*}{ DFGaze \cite{DFGAZE}} & \multirow{3}{*}{TIFS2024} & \multirow{3}{*}{Video-based} & 
        0.0 & 0.946 & 0.926 & 0.986 & 0.956 & 0.954 & 0.972 & 0.915 & 0.881 & 0.968 & 0.939& 0.921& 0.976
\\ & & & 
        0.4 & 0.854 & 0.724 & 0.795 & 0.756 & 0.726 & 0.824 & 0.818 & 0.709 & 0.743 & 0.810& 0.721& 0.788
\\ & & & 
        0.8 & 0.785 & 0.652 & 0.656 & 0.612 & 0.669 & 0.659 & 0.798 & 0.648 & 0.596 & 0.732& {\blue 0.657} & 0.637\\\hline
        \multirow{3}{*}{MaskRelation \cite{yang2023masked}} & \multirow{3}{*}{TIFS2023} & \multirow{3}{*}{Video-based} & 
        0.0 & 0.839&  0.838& 0.948
& 0.873& 0.749& 0.954
& 0.910& 0.814& 0.933
& 0.874& 0.800& 0.945
\\ & & & 
        0.4 & 0.815& {\blue 0.814} & {\blue 0.897}
& 0.811 & 0.543& {\blue 0.914}
& 0.868& 0.627& {\blue 0.906}
& 0.832& 0.662& {\blue 0.906}
\\ & & & 
        0.8 & 0.696& {\blue 0.696} & {\blue 0.692} & {\blue 0.790 }& 0.450 & {\blue 0.871}& 0.837& 0.456& {\blue 0.802} & 0.774& 0.534& {\blue 0.789} \\
        \hline         
        \multirow{3}{*}{LR-GCN [Ours]} & \multirow{3}{*}{-} & \multirow{3}{*}{Video-based} & 
        0.0 & {\red 0.962} & {\red 0.942} & {\red 0.989} & {\red 0.989} & {\red 0.968} & {\red 0.998} & {\red 0.969} & {\red 0.942} & {\red 0.988} & {\red 0.973} & {\red 0.951} & {\red 0.992}
\\ & & & 
        0.4 & {\red 0.958} & {\red 0.936} & {\red 0.987} & {\red 0.970} & {\red 0.920} & {\red 0.998} & {\red 0.969} & {\red 0.940} & {\red 0.988} & {\red 0.966} & {\red 0.932} & {\red 0.991}
\\ & & & 
        0.8 & {\red 0.944} & {\red 0.916} & {\red 0.983} & {\red 0.857} & {\red 0.738} & {\red 0.980} & {\red 0.962} & {\red 0.925} & {\red 0.979} & {\red 0.921} & {\red 0.860} & {\red 0.981}\\\hline
        \hline
    \end{tabular}\vspace*{-3mm}
    \label{tab:main}}\vspace*{-3mm}
\end{table*}

\begin{table}[ht]
\centering
\caption{Complexity comparison of different methods in terms of FLOPs , MACs, and \#Params.}
\label{tab:method_comparison}
\scalebox{1}{
\begin{tabular}{c|c|c|c}
\hline\hline
Method & FLOPs (G) & MACs (G) & \#Params (M) \\
\hline
Xception \cite{chollet2017xception} & 60.796 & 30.356 & 21.861 \\
$F^3$-net \cite{qian2020thinking} & 192.604 & 95.880 & 22.125 \\
RECCE \cite{RECCE} & 81.655 & 40.667 & 47.693 \\
CORE \cite{ni2022core} & 60.978 & 30.356 & 21.861 \\
UCF \cite{yan2023ucf} & 180.738 & 90.087 & 46.838 \\
DFIL \cite{pan2023dfil} & 60.976 & 30.356 & 20.811 \\
TALL-Swin \cite{xu2023tall} & 30.318 & 15.125 & 86.920 \\
DFGaze \cite{DFGAZE} & 22.647 & 11.241 & 123.185 \\
MaskRelation \cite{yang2023masked} & 46.665 & 23.315 & 28.4 \\
\hline
LR-GCN [Ours] & 70.751 & 35.246 & 29.661 \\
\hline
\hline
\end{tabular}}\vspace*{-3mm}\label{tab:complexity_analysis}\vspace*{-3mm}
\end{table}

The primary performance assessment comparing the handling of invalid facial images between our proposed model, LR-GCN, and various state-of-the-art schemes is provided in Table \ref{tab:main}. Under optimal conditions, where most facial images are valid, LR-GCN exhibits competitive results, holding its own against other cutting-edge DeepFake video detection methods, such as Xception \cite{chollet2017xception}, $F^3$-Net \cite{qian2020thinking}, RECCE \cite{RECCE}, CORE \cite{ni2022core}, DFIL \cite{pan2023dfil}, TALL-Swin \cite{xu2023tall}, DFGaze \cite{DFGAZE} and MaskRelation \cite{yang2023masked}. Note that TALL-Swin, DFGaze, and MaskRelation are video-based approaches.

Specifically, the F1-Score of LR-GCN 
slightly surpasses those of its contemporaries under clean cases (i.e., $m_r=0$). This outcome implies that the proposed LR-GCN with Graph Laplacian Spectral Prior Regularization is effective and reliable for DeepFake video detection. However, in scenarios where partial face images are invalid due to purposeful attacks on face detectors, the performance of traditional image-based methods, including Xception \cite{chollet2017xception},  $F^3$-Net \cite{qian2020thinking}, RECCE \cite{RECCE}, CORE \cite{ni2022core}, UCF \cite{yan2023ucf}, and DFIL \cite{pan2023dfil}, may substantially deteriorate since they fail to consider noisy face sequences in real-world scenarios.

Similarly, the video-level DeepFake detection methods, TALL-Swin \cite{xu2023tall} and DFGaze \cite{DFGAZE}, which heavily rely on temporal cues, may suffer further performance degradation when the masking ratio increases. Invalid faces can cause landmark detection failures and incorrect temporal trajectories. Consequently, the F1-score of TALL-Swin \cite{xu2023tall} and DFGaze \cite{DFGAZE} under a masking ratio of 0.8 in the testing phase is lower than 0.7, implying that all predictions would be categorized as either entirely fake or real. In stark contrast, all quality indices of our proposed LR-GCN, evaluated on different datasets, display promising results, suggesting that LR-GCN is robust and reliable even under highly noisy face sequences (e.g., when $m_r=0.8$). Remarkably, since most DeepFake detection methods fail to discuss the impact of noisy face sequences, the degraded performance is most likely predictable.

\subsection{Computational Complexity}
To further demonstrate the efficiency and practicality of the proposed method, we conduct a comprehensive complexity analysis and compare it with other state-of-the-art DeepFake detection methods. Table \ref{tab:complexity_analysis} presents the comparison results in terms of floating-point operations (FLOPs), multiply-accumulate operations (MACs), and the number of model parameters for each method with $16\times 3\times 144\times 144$ tensor for the fair comparison. It is evident that LR-GCN achieves a remarkable balance between computational complexity and performance. With 70.751 trillion FLOPs, 35.246 trillion MACs, and 29.661 million parameters, LR-GCN exhibits a moderate computational overhead compared to other methods, such as TALL-Swin \cite{xu2023tall}, DFGaze \cite{DFGAZE}, UCF \cite{yan2023ucf}, and RECCE \cite{RECCE}. Notably, LR-GCN outperforms these methods in terms of FLOPs and MACs while maintaining a comparable number of parameters. Moreover, LR-GCN demonstrates superior performance in handling noisy face sequences, as shown in the experimental results, despite having a similar complexity to methods like CORE \cite{ni2022core}, Xception \cite{chollet2017xception}, and DFIL \cite{pan2023dfil}. This highlights the effectiveness of the proposed LR-GCN in learning discriminative and robust representations for DeepFake detection. The complexity analysis further substantiates LR-GCN as a practical and efficient solution for real-world DeepFake detection challenges, offering a compelling trade-off between computational resources and detection accuracy.

\begin{figure}
		\centering
		\includegraphics[width=0.50\textwidth]{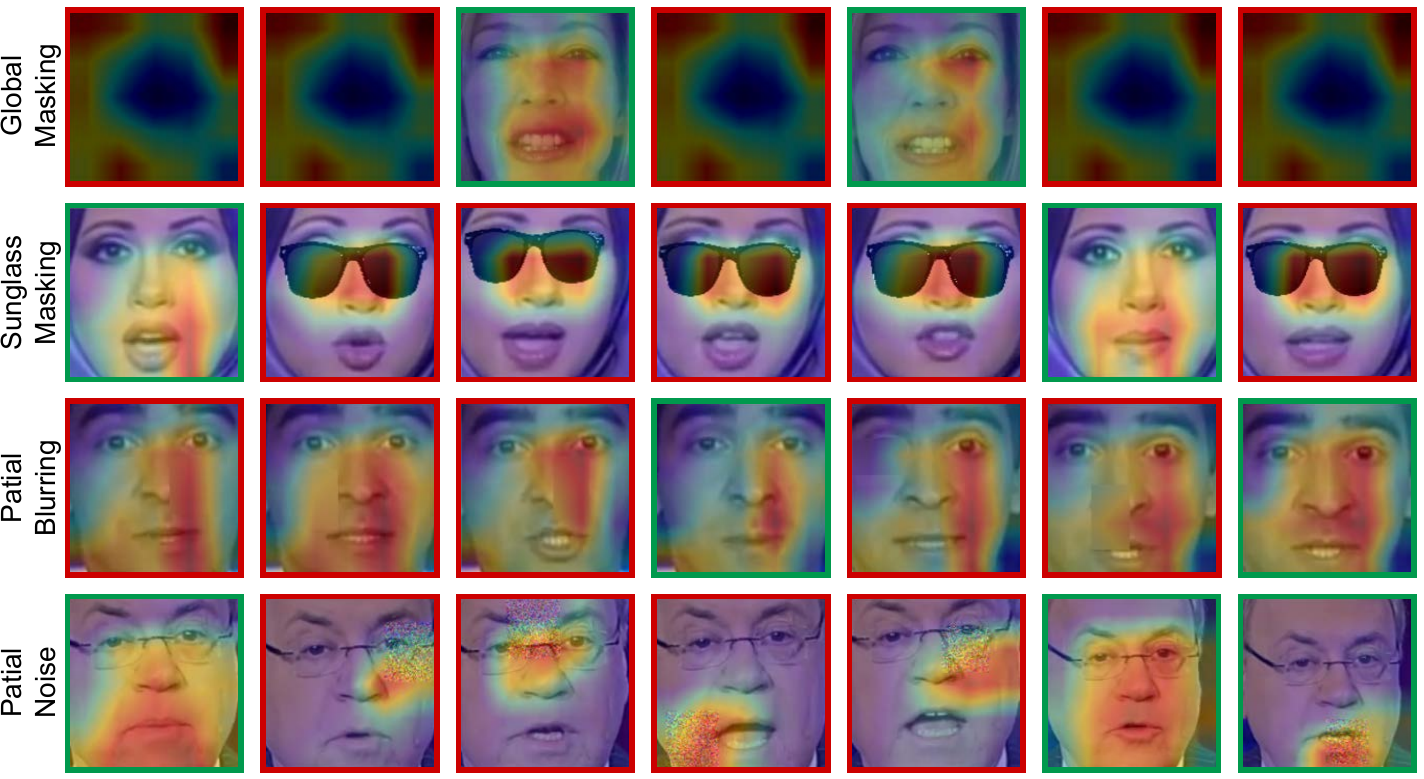}
		\caption{Examples of noisy face sequences with different perturbation types and Grad-CAM visualization \cite{gradcam}, showing frames labeled as either valid or masked, where masked frames represent various real-world corruptions: (i) global masking where faces are replaced with background, (ii) sunglasses masking with occlusions covering eye regions, (iii) partial blurring affecting specific facial areas, and (iv) partial noise in a specific patch of a random region.}\vspace*{-3mm}
		\label{fig:gradcam}	\vspace*{-3mm}
\end{figure}

\subsection{Robustness Evaluation under Perturbations}
As previously discussed, DeepFake videos can be intentionally perturbed to evade detection by face detectors, rendering traditional DeepFake detection methods ineffective. Beyond adversarial attacks, real-world scenarios often introduce additional perturbations, such as regional occlusions (e.g., sunglasses added mid-sequence, masks) or overlays (e.g., text, logos) that partially obscure facial features, as well as deliberate attempts to disrupt face detection. These distortions pose significant challenges to conventional methods, which often rely on consistent facial visibility or holistic feature extraction and struggle to adapt to such variability. To evaluate robustness under these conditions, we extend our experiments on the FF++ test set \cite{ffplus}, simulating both adversarial and real-world perturbations, with results reported in Table \ref{tab:realworld}.

For adversarial attacks, we employ an open-source PGD-like algorithm \cite{carlini2017towards} on the MTCNN face detector \cite{mtcnn}, using a maximum perturbation value of $\epsilon=0.04$, step size $\alpha_\text{adv}=0.01$, and $s=10$ iterations. In addition, to simulate real-world corruptions, we introduce regional occlusions by randomly adding sunglasses to the eye regions in a subset of frames, mimicking scenarios where occlusions appear dynamically (e.g., sunglasses worn mid-video). We also apply text overlays, logos, and random noise to portions of the frames, with masking ratios \( m_r = 0.2, 0.5, 0.8 \) representing the fraction of perturbed frames. These perturbations are designed to emulate practical challenges, such as partial facial obscurement or degraded image quality, as shown in Figure \ref{fig:gradcam}. Assuming each frame requires at least one detectable face, a black image replaces the frame if no face is detected. In the adversarial setting, an average of 3.58 faces are missed in the FF++ test set, akin to \( m_r = 0.2 \); however, the effective \( m_r \) may increase as adversarial noise can cause the detector to extract non-facial regions (e.g., background), further complicating detection.

Despite these challenges, LR-GCN demonstrates superior resilience compared to state-of-the-art methods, as shown in Table \ref{tab:realworld}.
While peer methods suffer significant performance drops due to missed detections, regional occlusions, and spatial distortions, LR-GCN maintains robust detection accuracy across all tested conditions. This advantage stems from the Order-Free Temporal Graph Embedding (OF-TGE), which constructs a local-temporal graph to dynamically select the most discriminative spatio-temporal cues. Unlike traditional approaches that depend on fixed temporal sequences or global facial features—leaving them vulnerable to regional or holistic perturbations—OF-TGE adaptively focuses on stable, informative patterns, bypassing occluded or noisy regions. This flexibility makes LR-GCN particularly well-suited to real-world scenarios, where both local (e.g., sunglasses, text overlays) and global (e.g., adversarial noise) distortions are prevalent.

The enhanced performance of LR-GCN under these diverse perturbations highlights its key contributions: (i) an adaptive, order-free graph structure that effectively handles both regional and global distortions, and (ii) a Laplacian-based spectral band-pass mechanism, jointly with dual-level sparsity constraints, that suppresses background semantics and random noise while preserving critical discriminative features. By addressing these practical challenges, LR-GCN fills a critical gap in existing DeepFake detection methods, offering a generalized, robust solution capable of tackling the complexities of real-world noisy face sequences. 
It is worth emphasizing that none of the perturbations in Table \ref{tab:realworld} (e.g., sunglasses masking, partial blurring/noise, or the PGD-like attack) is seen during training: LR-GCN is trained only on clean face sequences, and its robustness arises solely from the adaptive sparse graph representation, dual-level sparsity, and the explicit Laplacian-based prior, rather than corruption-specific data augmentation.

\begin{table*}
    \caption{The performance comparison of the proposed method and other methods trained on FF++ \cite{ffplus} under simulated real-world scenarios (i.e., local distortion and adversarial attack on face detector with $m_r^{local}=0.8$ and  $m_r^{adv}=0.2$ in our simulations, respectively).}
    \centering\small
    \scalebox{0.9}{
    \begin{tabular}{c|ccc|ccc|ccc|ccc|ccc}
    \hline\hline
    &  \multicolumn{3}{c|}{Sunglass Masking} &  \multicolumn{3}{c|}{Partial Blurring} & \multicolumn{3}{c|}{Partial Noisy} & \multicolumn{3}{c|}{Adversarial Attack} & \multicolumn{3}{c}{Average}    
    \\ \cline{2-16} 
   \multirow{-2}{*}{Method} & Acc. & F1 & AUC & Acc. & F1 & AUC & Acc. & F1 & AUC & Acc. & F1 & AUC & Acc. & F1 & AUC \\ \hline
{Xception \cite{chollet2017xception}} & 0.813& 0.594& 0.653& 0.813& 0.594& 0.653& 0.813& 0.594& 0.653& 0.814& 0.614& 0.585& 0.813& 0.599& 0.636
\\
{\text{$F^3$}-net \cite{qian2020thinking}} & 0.817& 0.598& 0.662& 0.817& 0.598& 0.662& 0.817& 0.598& 0.662& 0.764& 0.480& 0.504& 0.804& 0.569& 0.623
\\
{RECCE \cite{RECCE}} & 0.817& 0.599& 0.655& 0.817& 0.599& 0.655& 0.817& 0.599& 0.655& 0.794& 0.444& 0.527& 0.811& 0.560& 0.623
\\
{CORE \cite{ni2022core}} & 0.818& 0.601& 0.661& 0.818& 0.601& 0.661& 0.818& 0.601& 0.661& 0.793&  0.445& 0.503& 0.812& 0.562& 0.622
\\ 
{UCF \cite{yan2023ucf}} & 0.815& 0.598& 0.660& 0.815& 0.597& 0.660& 0.815& 0.597& 0.660& 0.727& 0.603& 0.586& 0.793& 0.599& 0.642\\

{DFIL \cite{pan2023dfil}} & 0.802 & 0.678 & 0.669 & 0.805 & 0.691 & 0.695 & 0.825 & 0.825 & 0.701 & 0.755 & 0.691 & 0.805 & 0.797 & 0.721 & 0.718
\\ 

{TALL-Swin \cite{xu2023tall}} & 0.815 & 0.648 & 0.615 & {\blue 0.826} & 0.658 & 0.648 & {\blue 0.846} & 0.636 & 0.621 & 0.796 & 0.715 & 0.701 & {\blue 0.821} & 0.664& 0.683
\\ 

{DFGaze \cite{DFGAZE}} & 0.801 & 0.579 & 0.649 & 0.814 & 0.679 & 0.684 & 0.819 & 0.648 & 0.685 & 0.707 & 0.628 & 0.746 & 0.785& 0.634& 0.691
\\
{MaskRelation \cite{yang2023masked}} & {\blue 0.819} & {\blue 0.786} & {\blue 0.956} & 0.823 & {\blue 0.822} & {\blue 0.830} & 0.706 & {\blue 0.704} & {\blue 0.849} & {\blue 0.805} & {\blue 0.791} & {\blue 0.876} & 0.788 & {\blue 0.775} & {\blue 0.877}\\\hline

LR-GCN (Ours) & {\color{red} 0.960} & {\color{red} 0.932} & {\color{red} 0.962} & {\color{red} 0.960} & {\color{red} 0.932} & {\color{red} 0.997} & {\color{red} 1.000} & {\color{red} 1.000} & {\color{red} 1.000} & {\color{red} 0.910} & {\color{red} 0.883} & {\color{red} 0.937} & {\color{red} 0.957} & {\color{red} 0.936} & {\color{red} 0.974} \\\hline\hline
    \end{tabular}}\vspace*{-2mm}\label{tab:realworld}\vspace*{-2mm}
\end{table*}

\subsection{Ablation Study}

Table \ref{tab:module} presents an ablation study for the proposed modules in our LR-GCN, \textit{i.e.}, GCN, GLSP, and SC, where the performance is evaluated in noisy face sequences (say, $m_r=0.8$ and $m_r=0.7$). When GLSP is enabled, we activate the Laplacian pre-filter in \eqref{eq:lap_pre} (implemented by the GCNGSP module), whereas the ``GCN only'' setting feeds $\mathbf{X}$ directly into the GCN layers without explicit Laplacian filtering.
 Note that when none of the proposed modules is adopted, we adopt the Transformer \cite{vit} as the classification head with four-head multi-head self-attention (MHSA) with the embedding size of 512 to meet a similar number of parameters as our LR-GCN, which could be treated as a variant of Convolutional Transformer. When we enable the GCN for the proposed adaptive affinity matrix and its affinity matrix, the performance of the DeepFake video detection under noisy face sequences, implying that the adaptive affinity matrix and its graph representation judiciously embeds the different spatiotemporal features into every node, thereby reducing the impact of invalid faces under noisy face sequences. Furthermore, the proposed GLSP could improve the robustness since it could filter noisy nodes containing many invalid faces without significantly increasing computational complexity. Moreover, the proposed SC encourages the sparse feature representation in $\mathbf{X}$, enforcing our ASGE to be even sparser, thereby significantly enhancing the importance of a few spatiotemporal features, which are exactly contributed by valid faces, and therefore, improving the robustness and performance for DeepFake video detection. In the appendix, we also provide experiments on hyperparameter selection for our LR-GCN. More ablation studies, including hyperparameter selection and cross-dataset experiments, are listed in the appendix.

 Although Table \ref{tab:module} primarily ablates the GCN, GLSP, and SC components, we emphasize that all variants rely on the same order-free ASGE. In contrast, a temporal-GCN baseline that connects nodes only along strictly ordered frame indices $$(t, t±1)$$ is inherently brittle under our masking protocol; in our preliminary experiments, it exhibits substantially larger performance drops as $m_r$ increases.  Future work could explore hybrid architectures that reintroduce lightweight temporal positional encodings when the input quality assessment deems the sequence reliable, thereby combining the robustness of order-free graphs with the discriminativeness of explicit temporal modeling.

\setlength{\tabcolsep}{4pt} 
\begin{table}
\caption{Ablation study of the proposed LR-GCN using different classification heads and components. 
}
\centering
\begin{tabular}{c|ccc|ccc|cc}
\hline\hline
\small
$m_r$ & GCN & GLSP & SC & Acc. & F1 & AUC & \multicolumn{2}{c}{ \#Param / MACs} \\ \hline 
0.8 & & & & 0.844 & 0.655 & 0.705 & \multicolumn{2}{c}{40.79M / 38.82} \\\hline
0.8 & \checkmark & & & 0.926 & 0.873 & 0.931 & \multicolumn{2}{c}{\multirow{4}{*}{29.66M / 35.25}} \\
0.8 & \checkmark & \checkmark & & 0.924 & 0.876 & {\blue 0.977} & & \\
0.8 & \checkmark & & \checkmark & {\red 0.946} & {\blue 0.912} & 0.975 & &\\ 
0.8 & \checkmark & \checkmark & \checkmark & {\blue 0.944} & {\red 0.916} & {\red 0.983}  & &\\ \hline\hline
0.7 & & & & 0.858 & 0.700 & 0.794 & \multicolumn{2}{c}{40.79M / 38.82} \\\hline
0.7 & \checkmark & & & 0.938 & 0.900 & 0.965 & \multicolumn{2}{c}{\multirow{4}{*}{29.66M / 35.25}} \\
0.7 & \checkmark & \checkmark & & 0.950 & 0.920 & {\blue 0.982} & &\\
0.7 & \checkmark & & \checkmark & {\blue 0.952} & {\blue 0.925} & 0.974 & &\\ 
0.7 & \checkmark & \checkmark & \checkmark & {\red 0.960} & {\red 0.938} & {\red 0.985} & &\\ \hline\hline
\end{tabular}\vspace*{-2mm}\label{tab:module}
\vspace*{-2mm}
\end{table}

\begin{figure}[!t]
\begin{center}
\scalebox{0.52}
{
\begin{tabular}[c]{c@{ }c@{ }}
\hspace{-6mm}
    \includegraphics[width=0.47\textwidth,valign=t]{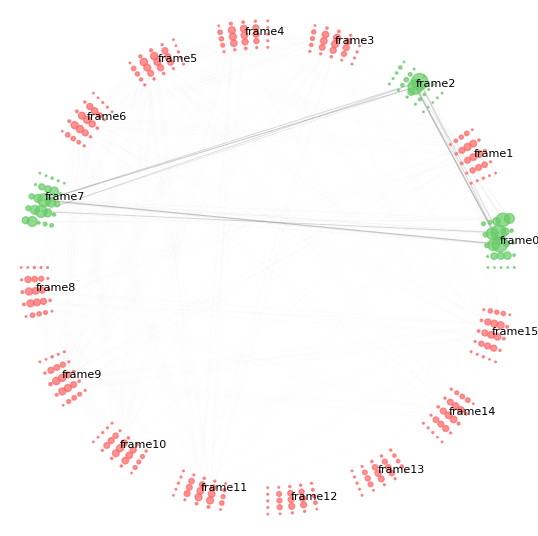}&
  	\includegraphics[width=0.47\textwidth,valign=t]{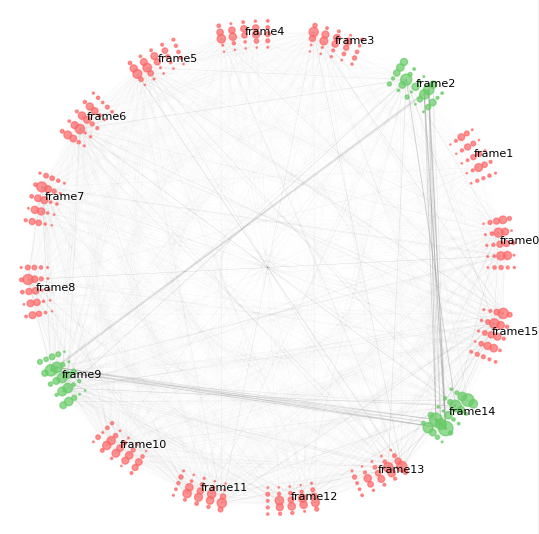}
    \\
    
    \hspace{-6mm}
    \Large~Face2Face\_506\_478\_partial\_noisy & \Large~Deepfakes\_285\_136\_partial\_blurring \\
\hspace{-6mm}
    \includegraphics[width=0.47\textwidth,valign=t]{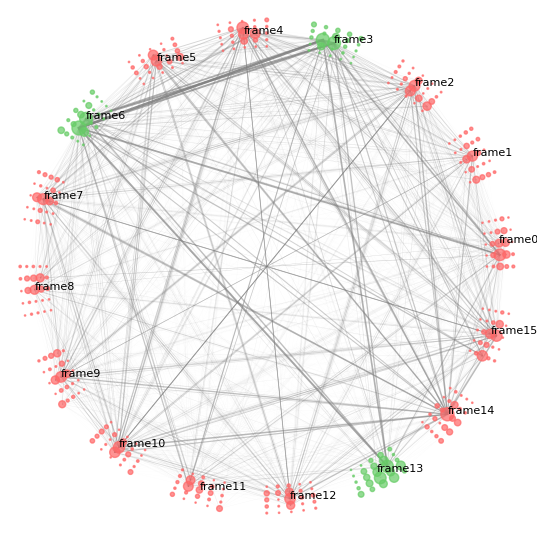}&
  	\includegraphics[width=0.47\textwidth,valign=t]{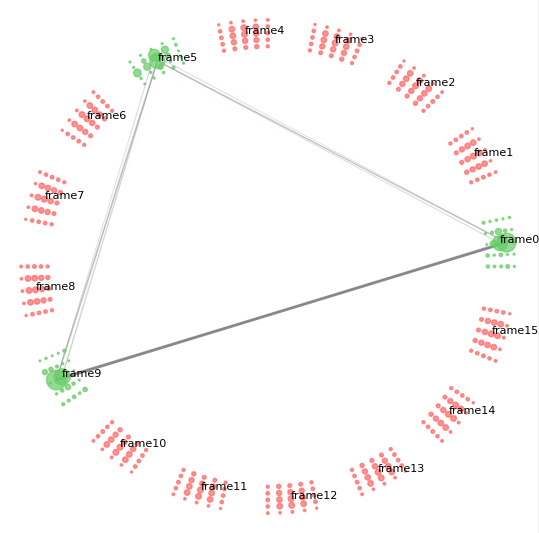}
    \\
    
    \hspace{-6mm}
    \Large~FaceSwap\_375\_251\_sunglass\_masking & \Large~NeuralTextures\_842\_714\_global\_missing \\\hspace{-6mm}
    \includegraphics[width=0.47\textwidth,valign=t]{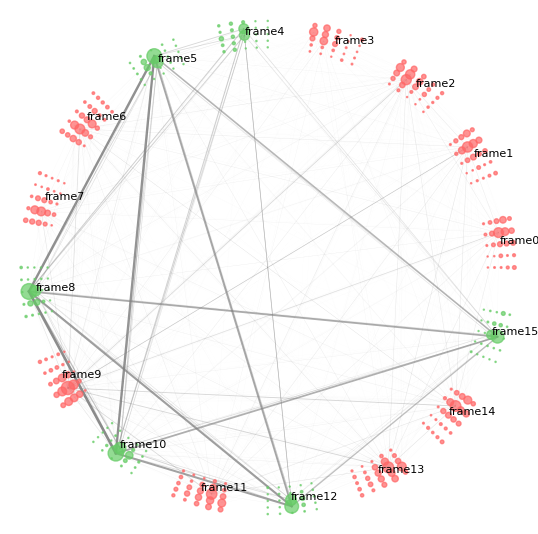}&
  	\includegraphics[width=0.47\textwidth,valign=t]{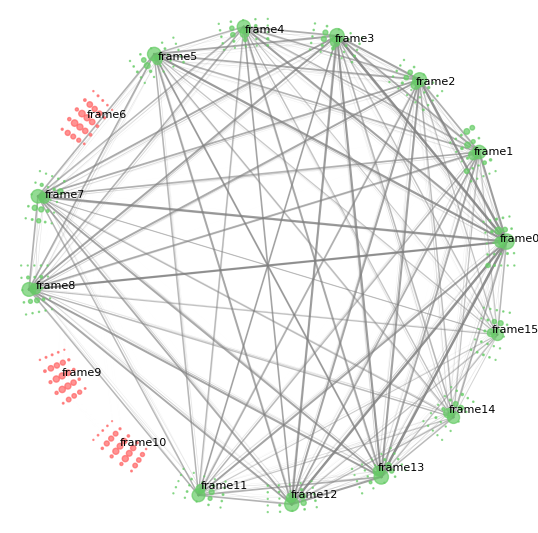}
    \\
    
    \hspace{-6mm}
    \Large~Real\_youtube\_885\_partial\_blurring & \Large~Real\_youtube\_129\_partial\_noisy \\
    
\end{tabular}
}
\end{center}
\vspace{-4mm}
\caption{{Graph visualization across different mask types and masking ratios for FF++\cite{ffplus}, illustrating the adaptive sparse graphs constructed for various perturbation scenarios. Green nodes represent valid facial features, while red nodes indicate corrupted or masked regions. Edge connections show how the proposed approach adaptively maintains meaningful relationships between valid nodes despite significant corruptions.}}
\label{fig:graph_feature}\vspace{-4mm}
\end{figure}

\subsection{Feature Analysis}

An in-depth visual analysis provides critical insights into the behavior and robustness of DeepFake detection methods under complex real-world perturbation scenarios, such as adversarial attacks, regional occlusions, and unstable face detection processes.

Figure \ref{fig:graph_feature} presents an explicit visualization of the adaptive sparse graphs generated by the proposed ASGE under various realistic manipulation scenarios from the FF++ dataset \cite{ffplus}. Specifically, valid nodes (marked green) consistently form distinct and coherent subgraphs, maintaining stable connections even under significant corruption. Conversely, corrupted nodes (marked red), representing perturbed or occluded regions, exhibit fewer and weaker connections, effectively isolating noise and preventing erroneous feature propagation. This clear separation demonstrates the capability of the proposed OF-TGE and LR-GCN to adaptively filter noisy information at both structural and feature levels, preserving meaningful and discriminative relationships essential for accurate classification.



Overall, these comprehensive visual analyses provide strong empirical support for the proposed ASGE, OF-TGE, and LR-GCN methodologies, demonstrating their collective ability to robustly address complex perturbations typical in practical DeepFake detection scenarios.

\subsection{Limitations and Discussion}

This study introduces a novel approach, LR-GCN, to address the challenge of DeepFake video detection in the presence of noisy face sequences. LR-GCN leverages an adaptive affinity matrix with sparse constraints and a graph convolutional network equipped with a Graph Laplacian Spectral Prior, yielding a task-driven spectral band-pass behavior that effectively exploits spatiotemporal correlations in face sequences while suppressing the impact of noise and distortions. The experimental results demonstrate the efficacy of LR-GCN in handling noisy face sequences and achieving state-of-the-art performance on several benchmark datasets.

One limitation is the trade-off inherent in the order-free design. By discarding strict temporal ordering to gain robustness against frame dropping and shuffling, the model may become less sensitive to forgeries that rely primarily on long-range temporal sequence anomalies (e.g., rhythmic unnaturalness or audio-visual synchronization). 
However, our experiments suggest that for current state-of-the-art DeepFakes, spatial and local-temporal inconsistencies captured by the affinity graph remain highly discriminative. 
Furthermore, the proposed spectral band-pass mechanism effectively compensates for the lack of explicit temporal order by identifying structural anomalies in the feature space.
Future work could explore hybrid architectures that re-integrate temporal positional encodings when the input quality assessment deems the sequence reliable.

Another aspect to consider is that LR-GCN currently does not incorporate masked learning strategies, which have shown promise in handling occlusions and missing data in various computer vision tasks. Integrating masked learning techniques into the LR-GCN framework could potentially further enhance its robustness to partial occlusions and incomplete face sequences. Moreover, the use of graph convolutional networks in LR-GCN allows for flexible processing of video frames, as the input frames are not required to be strictly sequential. This property could be leveraged to develop more efficient and adaptive sampling strategies for processing long video sequences.

It is also worth noting that while LR-GCN has demonstrated significant improvements over existing methods, there is still room for further enhancements. One direction could be to explore more advanced graph neural network architectures, such as graph attention networks or graph transformers, to better capture the complex dependencies and interactions among the spatiotemporal features. Additionally, incorporating prior knowledge or constraints specific to the DeepFake detection domain, such as the consistency of facial landmarks or the coherence of audio-visual signals, could potentially boost the performance and generalizability of the proposed approach.

In summary, LR-GCN represents a significant step forward in addressing the challenge of DeepFake video detection in the presence of noisy face sequences. While acknowledging the limitations and potential areas for improvement, we believe that the proposed methodology opens up new avenues for research in this critical domain. Future work could focus on extending LR-GCN to handle cross-dataset scenarios, integrating masked learning techniques, exploring more advanced graph neural network architectures, and incorporating domain-specific prior knowledge. As DeepFake techniques continue to evolve and become more sophisticated, developing robust and reliable detection methods that can operate effectively in real-world scenarios with noisy and challenging data remains an ongoing research endeavor of paramount importance.

\section{Conclusions}\label{sec:conclusions}
In this work, we presented LR-GCN, a robust DeepFake detection framework designed to handle unstable and corrupted facial sequences. By decoupling from temporal order via Order-Free Temporal Graph Embedding and enforcing dual-level sparsity alongside an explicit Graph Laplacian Spectral Prior that induces a spectral band-pass behavior, our method effectively isolates noise while preserving critical discriminative features. Extensive evaluations across standard benchmarks validate its superiority over state-of-the-art approaches, particularly under high perturbation settings. Future work may explore cross-domain generalization, integration with audio-visual signals, and adaptation to long-form video contexts to further enhance deployment readiness in unconstrained environments.

\begin{figure*}
    \centering
    \subfigure[Comparison of AUC for FF++]{
        \label{fig:result_FF++}
        \includegraphics[width=0.32\textwidth]{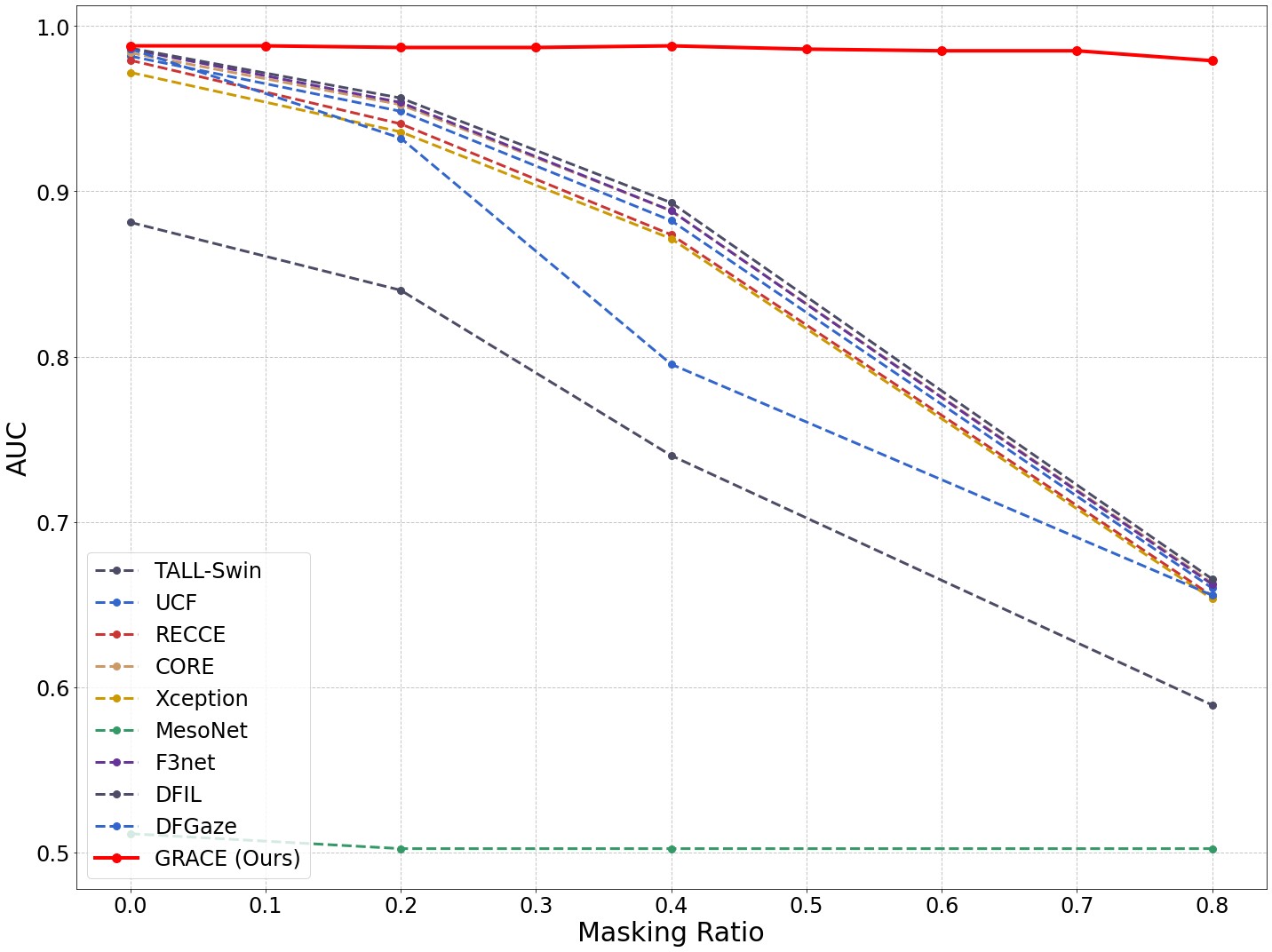}}
    \subfigure[Comparison of AUC for DFDC]{
        \label{fig:result_DFDC}
        \includegraphics[width=0.32\textwidth]{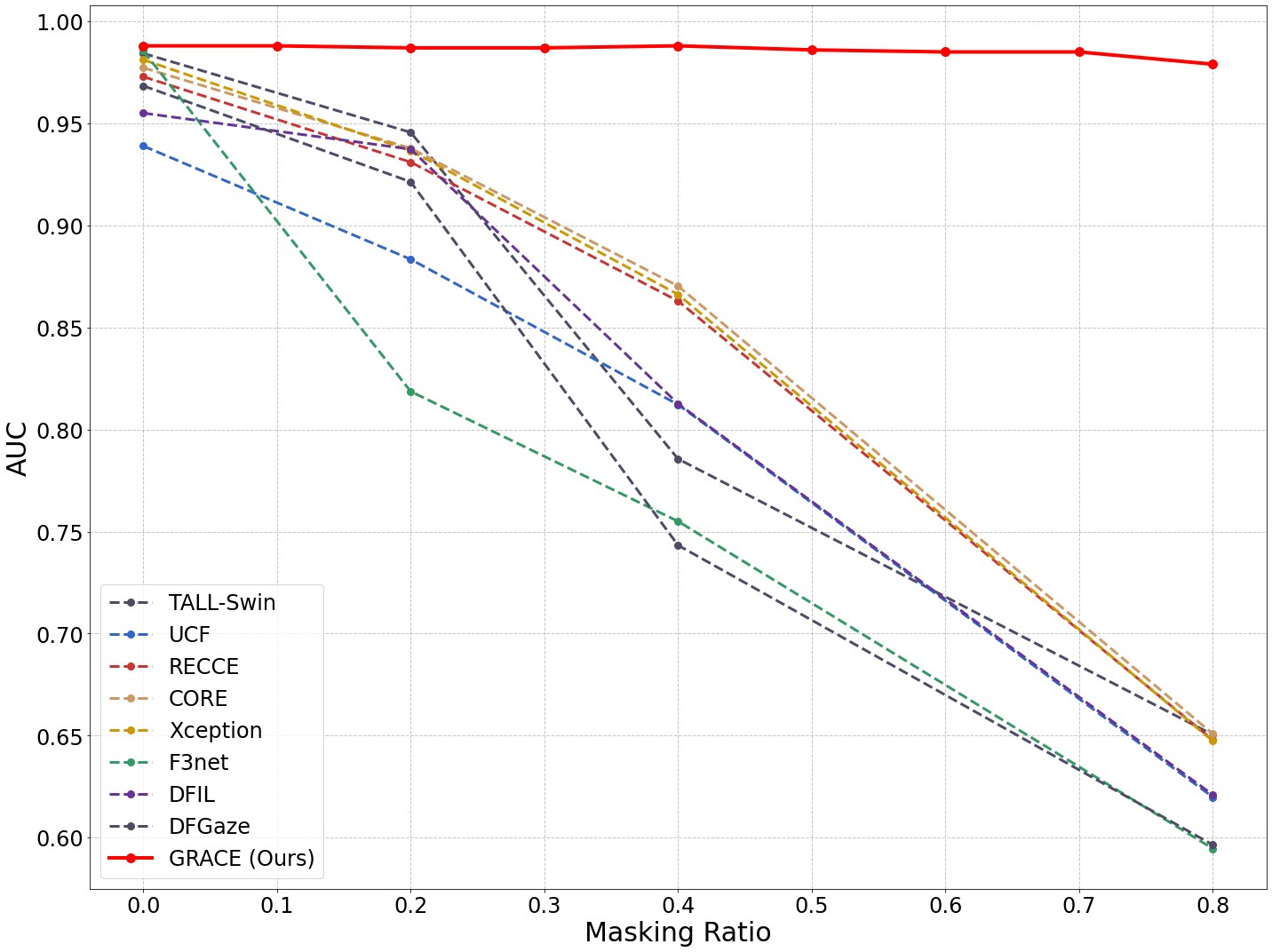}}
    \subfigure[Comparison of AUC for CelebDF]{
        \label{fig:result_celeb}
        \includegraphics[width=0.32\textwidth]{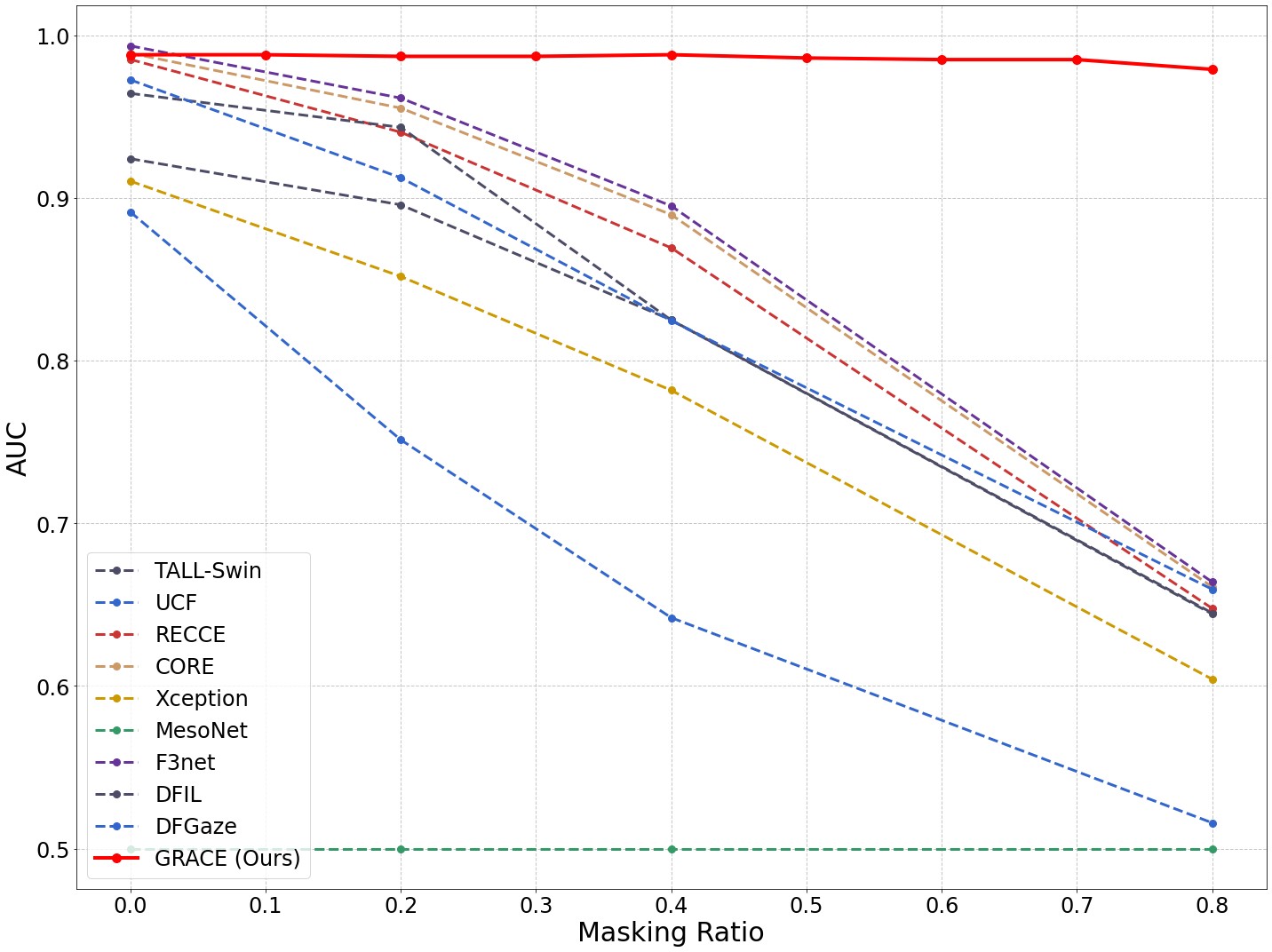}}
    \caption{{ The performance comparison of the proposed LR-GCN and other state-of-the-art methods in terms of AUC under different masking ratios $m_r$ for (a) FF++\cite{ffplus}, (b) DFDC \cite{dfdc}, and (c) Celeb-DF \cite{celeb}.}}
    \label{Fig.main}
\end{figure*}

\begin{figure}
		\centering
		\includegraphics[width=0.47\textwidth]{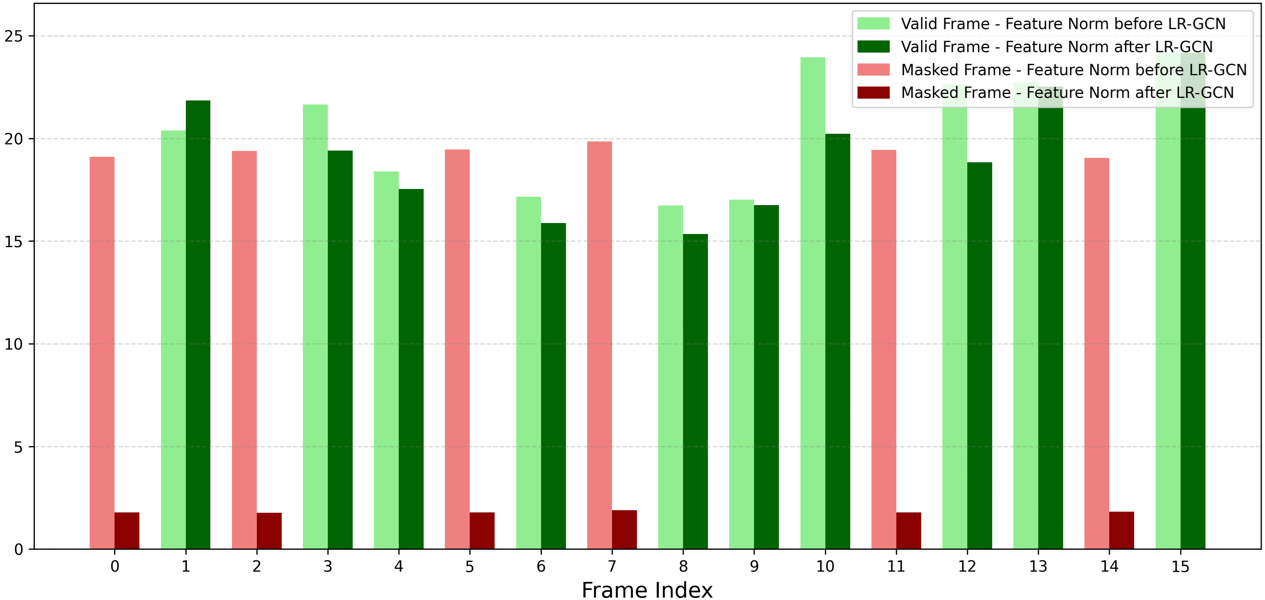}
		\caption{{{The illustration of feature magnitude distributions across frames, showing valid frames (green) with high feature values and invalid frames (red) with partial noise before/after the proposed LR-GCN.}}}
		\label{fig:oftge}	
\end{figure}

\section*{Additional Quantitative Results Analysis}

The detailed quantitative results, evaluated on benchmark datasets, are illustrated in Figures \ref{fig:result_FF++}, \ref{fig:result_DFDC} and \ref{fig:result_celeb}. In the clean case, i.e., when $m_r=0$, the performance of the proposed method is comparable to other state-of-the-art methods. It is observed that performance degradation becomes increasingly pronounced with a rise in the masking ratio during the testing phase, particularly when the masking ratio ($m_r$) exceeds 0.5. The performance of the previously established TALL-Swin \cite{xu2023tall} and DFGaze \cite{DFGAZE} also decline when the masking ratio surpasses 0.2. A similar trend is discernible in Figure \ref{fig:result_celeb}, which evaluates the DFDC testing set. The performance of contemporary methods diminishes at higher masking ratios, whereas the proposed LR-GCN method maintains relatively high performance even at a masking ratio of 0.8.

More specifically, most existing DeepFake video/image detection algorithms do not address the impact of noisy face sequences. Although state-of-the-art face detectors perform exceptionally well under pristine conditions, their performance can be severely undermined when subjected to well-engineered post-processing techniques, particularly adversarial perturbations targeting the face detector. Our LR-GCN method successfully overcomes this shortcoming and introduces a novel and robust DeepFake video detection approach for real-world challenges.

\section*{Invalid Node Filtering Mechanism}
To elucidate the effectiveness of our proposed Laplacian-Regularized Graph Convolutional Network (LR-GCN), we analyze the feature norm magnitudes across frames in noisy face sequences, as illustrated in Fig. \ref{fig:oftge}. This figure visualizes the distributions of feature norms for both valid and invalid frames before and after applying LR-GCN. Specifically, the x-axis represents the frame index (ranging from 0 to 15), while the y-axis indicates the feature norm magnitude (scaled from 0 to 25). Four categories are depicted: valid frames before LR-GCN (light green), valid frames after LR-GCN (dark green), invalid frames before LR-GCN (light red), and invalid frames after LR-GCN (dark red).

The results highlight LR-GCN’s capability to enhance valid facial information while suppressing the influence of invalid faces in noisy sequences. For valid frames that carry authentic facial features, the feature norm magnitudes increase and stabilize post-processing, rising from fluctuating values (e.g., peaking at 15–20) to consistently higher levels (approximately 20 or above). This amplification underscores LR-GCN’s ability to strengthen discriminative features critical for accurate detection tasks, such as identifying DeepFake manipulations.

In contrast, for invalid frames—those degraded by noise, occlusions, or other distortions—LR-GCN effectively suppresses their feature norm magnitudes. Initially, these frames exhibit norms ranging from 15 to 20, retaining residual signals that could interfere with model performance. After applying LR-GCN, their norms drop significantly to near-zero values (approximately 2–3), demonstrating that LR-GCN filters out noise and irrelevant information. This suppression is pivotal in ensuring that invalid faces do not mislead the detection process, particularly in noisy face sequences where corrupted inputs are prevalent.

The mechanism behind this behavior lies in LR-GCN’s Graph Laplacian Spectral Prior (GLSP) combined with GCN aggregation. The normalized Laplacian first acts as a high-pass operator that emphasizes discrepancies between semantically related nodes, producing residual responses that are particularly sensitive to structural artifacts and detector failures. The subsequent GCN layers then perform low-pass aggregation over these residuals: isolated spikes corresponding
to random noise are smoothed out due to lack of neighbor support, whereas consistent responses along manipulated or occluded regions are reinforced. Consequently, invalid frames —whose activations are noisy and inconsistent across the graph—are effectively suppressed, while valid frames retain strong, stable embeddings that drive reliable DeepFake detection. This spectral band-pass behavior enhances robustness against real-world challenges, such as incomplete or degraded face sequences, and mitigates the impact of invalid faces in noisy face sequences.

\subsection*{Hyperparameters Selection}

\begin{table}
\caption{Performance evaluation of the proposed LR-GCN with different hyperparameter settings using FF++ \cite{ffplus}. $g_\text{dim}$ and $g_n$ are the embedding dimension and number of layers of GCN, respectively; $N$ is the frames extracted from the video; $n_\text{out}$ is the number of neurons of FC; $\lambda$ stands for weights of sparsity; $\lambda$ is the weight of regularization term. The highest AUC scores in each section are highlighted in \textcolor{red}{red}.}
\centering
\begin{tabular}{c | c c | c c | c c | c c}
\hline \hline
$m_r$ & $N$ & AUC & $g_n$ & AUC & $\lambda$ & AUC & $g_\text{dim}$ & AUC \\
\hline \hline
0.8 & 12 & 0.971 & 12 & 0.950 & 1e$^{-7}$ & 0.978 & 600 & 0.966 \\
0.8 & 20 & 0.978 & 4 & 0.982 & 1e$^{-6}$ & 0.974 & 200 & 0.981 \\
0.8 & 16 & \textcolor{red}{0.983} & 8 & \textcolor{red}{0.983} & 1e$^{-5}$ & \textcolor{red}{0.983} & 400 & \textcolor{red}{0.983} \\
\hline
0.7 & 12 & 0.983 & 12 & 0.964 & 1e$^{-7}$ & 0.981 & 600 & 0.984 \\
0.7 & 20 & \textcolor{red}{0.986} & 4 & 0.983 & 1e$^{-6}$ & 0.979 & 200 & \textcolor{red}{0.987} \\
0.7 & 16 & 0.985 & 8 & \textcolor{red}{0.985} & 1e$^{-5}$ & \textcolor{red}{0.985} & 400 & 0.985 \\
\hline
\multicolumn{9}{c}{} \\ [-1.5ex]
\hline \hline
$m_r$ & \multicolumn{2}{c|}{$n_\text{out}$} & \multicolumn{2}{c|}{AUC} & \multicolumn{2}{c|}{$\beta$} & \multicolumn{2}{c}{AUC} \\
\hline \hline
0.8 & \multicolumn{2}{c|}{1024} & \multicolumn{2}{c|}{0.969} & \multicolumn{2}{c|}{0.1} & \multicolumn{2}{c}{0.943} \\
0.8 & \multicolumn{2}{c|}{3072} & \multicolumn{2}{c|}{0.939} & \multicolumn{2}{c|}{0.5} & \multicolumn{2}{c}{\textcolor{red}{0.983}} \\
0.8 & \multicolumn{2}{c|}{2048} & \multicolumn{2}{c|}{\textcolor{red}{0.983}} & \multicolumn{2}{c|}{1.0} & \multicolumn{2}{c}{0.940} \\
\hline
0.7 & \multicolumn{2}{c|}{1024} & \multicolumn{2}{c|}{0.984} & \multicolumn{2}{c|}{0.1} & \multicolumn{2}{c}{0.960} \\
0.7 & \multicolumn{2}{c|}{3072} & \multicolumn{2}{c|}{0.968} & \multicolumn{2}{c|}{0.5} & \multicolumn{2}{c}{\textcolor{red}{0.985}} \\
0.7 & \multicolumn{2}{c|}{2048} & \multicolumn{2}{c|}{\textcolor{red}{0.985}} & \multicolumn{2}{c|}{1.0} & \multicolumn{2}{c}{0.965} \\
\hline \hline
\end{tabular}
\label{tab:hp_select}
\end{table}

\begin{table}[ht]
\centering
\small
\caption{Cross-dataset AUC (\%) on FF++ (C23) as held-in, and testing on Celeb-DF / DFDC under clean (no masking) condition. The last column $\Delta_C$ and $\Delta_D$ indicate the AUC drop from FF++ to Celeb-DF and DFDC. }
\label{tab:cross_clean_decimal}
\scalebox{0.95}{
\begin{tabular}{l|c|c|c|cc}
\hline
\textbf{Method} & \textbf{FF++} & \textbf{Celeb-DF} & \textbf{DFDC} & \(\Delta_C\) & \(\Delta_D\) \\
\hline
Xception        
& 0.972 & 0.737 & 0.709 
& 0.235 & 0.263 \\

\textit{F}$^3$\textit{-net}    
& \textcolor{deepgreen}{0.986} & 0.757 & 0.612 
& 0.229 & 0.374 \\

RECCE           
& 0.979 & 0.687 & -     
& 0.292 & -     \\

CORE            
& 0.984 & 0.794 & 0.757 
& 0.190 & 0.227 \\

UCF             
& 0.982 & 0.824 & \textcolor{blue}{0.805}
& 0.158 & \textcolor{deepgreen}{0.177} \\

DFIL            
& \textcolor{blue}{0.987} & 0.665 & 0.640 
& 0.322 & 0.347 \\

TALL-Swin       
& 0.881 & \textcolor{blue}{0.819} & 0.748
& \textcolor{blue}{0.062} & \textcolor{blue}{0.133} \\

DFGaze          
& \textcolor{deepgreen}{0.986} & 0.678 & 0.652
& 0.308 & 0.334 \\

MaskRelation    
& 0.973 & \textcolor{red}{0.950} & \textcolor{red}{0.902}
& \textcolor{red}{0.023} & \textcolor{red}{0.071} \\\hline

LR-GCN [Ours]   
& \textcolor{red}{0.989} & \textcolor{deepgreen}{0.906} & \textcolor{deepgreen}{0.769}
& \textcolor{deepgreen}{0.083} & 0.220 \\
\hline
\end{tabular}
}
\end{table}

\begin{table}[htbp]
    \centering
    \caption{LR-GCN Performance under Varying Masking Ratios. Held-in (FF++) and Cross-Dataset (Celeb-DF) AUC values are reported; Gain = Held-in AUC $-$$ $Celeb-DF AUC.}
    \scalebox{1}{
    \begin{tabular}{cccc}
        \toprule
        Masked Ratio $m_r$ & FF++(c23) & Celeb-DF & Drop \\
        \midrule
        0.0 & 0.989 & 0.9064 & 0.0826 \\
        0.1 & 0.989 & 0.8860 & 0.1030 \\
        0.2 & 0.990 & 0.8538 & 0.1362 \\
        0.3 & 0.987 & 0.9064 & 0.0806 \\
        0.4 & 0.987 & 0.8889 & 0.0981 \\
        0.5 & 0.986 & 0.9094 & 0.0766 \\
        0.6 & 0.988 & 0.8655 & 0.1225 \\
        0.7 & 0.985 & 0.8596 & 0.1254 \\
        0.8 & 0.983 & 0.7398 & 0.2432 \\
        \bottomrule
    \end{tabular}
    }
    \label{tab:masking}
\end{table}

To achieve optimal performance and robustness, we conducted a comprehensive ablation study to investigate the impact of various hyperparameters on the proposed LR-GCN method. This analysis provides valuable insights into the design choices and trade-offs involved in developing an effective DeepFake video detection system for real-world scenarios with noisy face sequences. Table \ref{tab:hp_select} presents the performance comparison of LR-GCN under different hyperparameter settings, evaluated on the FF++ dataset \cite{ffplus}.

\subsubsection*{1-1: Number of Extracted Frames ($N$)}
The number of frames employed during the training and testing phases is a crucial aspect of LR-GCN. While using a larger number of frames might intuitively improve performance, it also significantly increases the computational complexity. To strike an optimal balance, we investigated the impact of varying the number of extracted frames. As shown in Table \ref{tab:hp_select}, using $N=8$ frames results in the lowest computational complexity but slightly compromises performance in terms of AUC. Conversely, increasing the number of frames to $N=20$ achieves state-of-the-art performance for most masking ratios during testing. Considering the trade-off between effectiveness and efficiency, we recommend using $N=16$ frames as the optimal setting for LR-GCN.

\subsubsection*{1-2: Number of GCN Layers ($g_n$)}
The depth of the Graph Convolutional Network (GCN) plays a vital role in learning robust feature representations. However, stacking too many layers with the Graph Laplacian smooth prior may lead to over-smoothing of nodes and reduce the discriminative power. We explored the impact of varying the number of GCN layers ($g_n$) in LR-GCN. As presented in Table \ref{tab:hp_select}, setting $g_n=12$ results in suboptimal performance compared to $g_n=8$ and $g_n=4$, likely due to convergence difficulties within the given 200 epochs. While $g_n=4$ achieves outstanding performance overall, it slightly underperforms in highly noisy conditions (i.e., $m_r=0.8$) compared to $g_n=8$. Therefore, we suggest using $g_n=8$ as a balanced choice for stable and robust performance across various noise levels.

\subsubsection*{1-3: Sparsity Penalty Term ($\lambda$)}
The sparsity penalty term $\lambda$ in the proposed loss function controls the balance between the sparsity constraint and the classification objective. A higher value of $\lambda$ encourages LR-GCN to learn a sparser feature representation, which is particularly beneficial for DeepFake video detection in the presence of invalid facial images. We investigated the impact of $\lambda$ by varying its value from $1e^{-7}$ to $1e^{-5}$. As shown in Table \ref{tab:hp_select}, a higher sparsity penalty enhances the network's ability to learn essential and discriminative features, thereby reducing the influence of invalid faces and improving overall performance. However, setting $\lambda$ higher than $1e^{-5}$ leads to convergence difficulties. Based on our analysis, we recommend using $\lambda=1e^{-5}$ to achieve a balanced trade-off between sparsity and convergence stability.

\subsubsection*{1-4: GCN Embedding Dimension ($g_\text{dim}$)}
The embedding dimension of the GCN ($g_\text{dim}$) determines the richness of the learned feature representations for DeepFake video detection. We investigated the impact of $g_\text{dim}$ by comparing the performance of LR-GCN with $g_\text{dim} \in {200, 400, 600}$, as shown in Table \ref{tab:hp_select}. Since the dimension of the graph representation $\bA$ is $400\times 400$, intuitively, the best performance is achieved when $g_\text{dim}=400$. Reducing $g_\text{dim}$ below this value limits the expressive power of the GCN, while increasing it beyond introduces redundancy and harms performance. Therefore, we suggest setting $g_\text{dim}=400$ for optimal results.

\subsubsection*{1-5: Number of Fully Connected Layer Neurons ($n_\text{out}$)}
To aggregate the output of the GCN and feed it into the softmax classifier, a simple fully connected (FC) layer is employed, projecting the graph representation to an $n_\text{out}$-dimensional feature vector. We investigated the impact of $n_\text{out}$ by comparing the performance of LR-GCN with $n_\text{out} \in {1024, 2048, 3072}$, as shown in Table \ref{tab:hp_select}. While $n_\text{out}=2048$ achieves excellent performance under highly noisy face sequences, the performance gap between $n_\text{out}=2048$ and $n_\text{out}=1024$ is insignificant, suggesting that the choice of $n_\text{out}$ is not highly sensitive. Based on our analysis, we recommend setting $n_\text{out}=2048$ for a good balance between performance and computational complexity.

\subsubsection*{1-6: Threshold in Graph Construction ($\beta$)}

The threshold value $\beta$ determines the sparsity level of the constructed graph embedding. A smaller $\beta$ results in more edges in the graph, allowing more node features to be used for DeepFake video identification. However, this increases the model’s sensitivity to noisy face sequences, which may lead to minor performance degradation. Our experiments show that $\beta = 0.5$ achieves a good balance, providing strong performance and justifying its use as the default value.

The comprehensive analysis of the hyperparameters presented in this section highlights the robustness and effectiveness of the proposed LR-GCN method under various hyperparameter settings. By carefully selecting these hyperparameters, LR-GCN achieves state-of-the-art performance in DeepFake video detection, even in challenging real-world scenarios with noisy face sequences. The insights gained from this analysis provide valuable guidance for practitioners and researchers aiming to develop robust and efficient DeepFake detection systems.

\subsection*{Cross-Dataset Evaluation}
To evaluate the generalization capability of the proposed LR-GCN, we performed cross-dataset experiments by training the model on the FF++ dataset and testing it on the Celeb-DF and DFDC datasets. In the clean condition (i.e., without masking), LR-GCN achieves an AUC of 0.989 on the held-in FF++ dataset, outperforming all compared methods. When tested on the Celeb-DF dataset, LR-GCN records an AUC of 0.906, securing the second-best performance behind MaskRelation (0.950). Notably, the performance drop, defined as \(\Delta_C = 0.083\), is substantially smaller than that of competing approaches, such as Xception (0.235) and DFIL (0.322), highlighting LR-GCN's robustness to domain shifts—an essential property for detecting DeepFake videos across diverse sources. Although the proposed LR-GCN is not designed to improve the robustness of cross-dataset scenarios, the performance remains strong compared to existing state-of-the-art methods.

Additionally, we examined the effect of varying masking ratios on cross-dataset performance. Remarkably, even at a high masking ratio (e.g., \(m_r = 0.5\)), LR-GCN sustains an AUC of 0.9094 on Celeb-DF, with a modest gain of 0.0766. This resilience underscores the efficacy of the proposed order-free graph embedding and dual-level sparsity constraints, which collectively mitigate the detrimental impact of unstable facial sequences, thereby enhancing the model’s generalization across datasets and under noisy conditions.

These results affirm the core strengths of LR-GCN, demonstrating its robustness in addressing domain shifts and noisy inputs, both of which are critical challenges in real-world DeepFake detection scenarios.

	\bibliographystyle{IEEEtran}
	\bibliography{sn-bibliography}

\vspace{-13mm}
\begin{IEEEbiography}[{\includegraphics[width=1in,height=1in,clip,keepaspectratio]{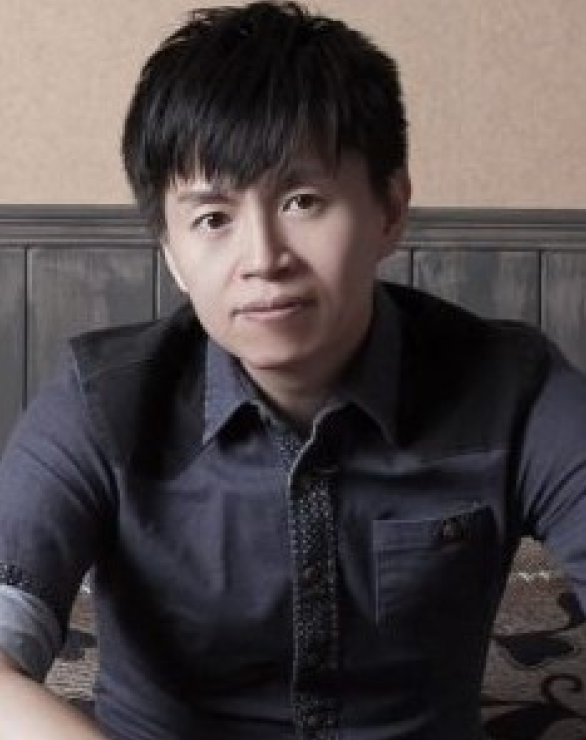}}]%
{Chih-Chung Hsu}
(Senior Member, IEEE) received the Ph.D. degree in Electrical Engineering from National Tsing Hua University, Taiwan, in 2014. He served as an Assistant Professor at National Pingtung University of Science and Technology (2018–2021) and later at National Cheng Kung University (NCKU). Since 2025, he has been an Associate Professor with the Institute of Intelligent Systems, National Yang Ming Chiao Tung University (NYCU), Taiwan. His research interests include computer vision, deep learning, and multimedia forensics. Dr. Hsu has published in IEEE TPAMI, TIP, and CVPR, and received the Best Student Paper Award at ICIP 2019 along with multiple international challenge championships.
\end{IEEEbiography}

\vspace{-10mm}

\begin{IEEEbiography}[{\includegraphics[width=1in,height=1in,clip,keepaspectratio]{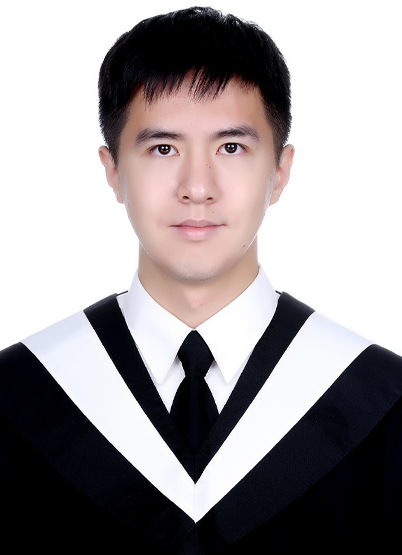}}]%
{Shao-Ning Chen} received the M.S. degree in Data Science from National Cheng Kung University (NCKU), Taiwan, in 2022. His research focuses on image processing and computer vision, specifically video DeepFake detection in unreliable face sequences.
\end{IEEEbiography}

\vspace{-10mm}

\begin{IEEEbiography}[{\includegraphics[width=1in,height=1in,clip,keepaspectratio]{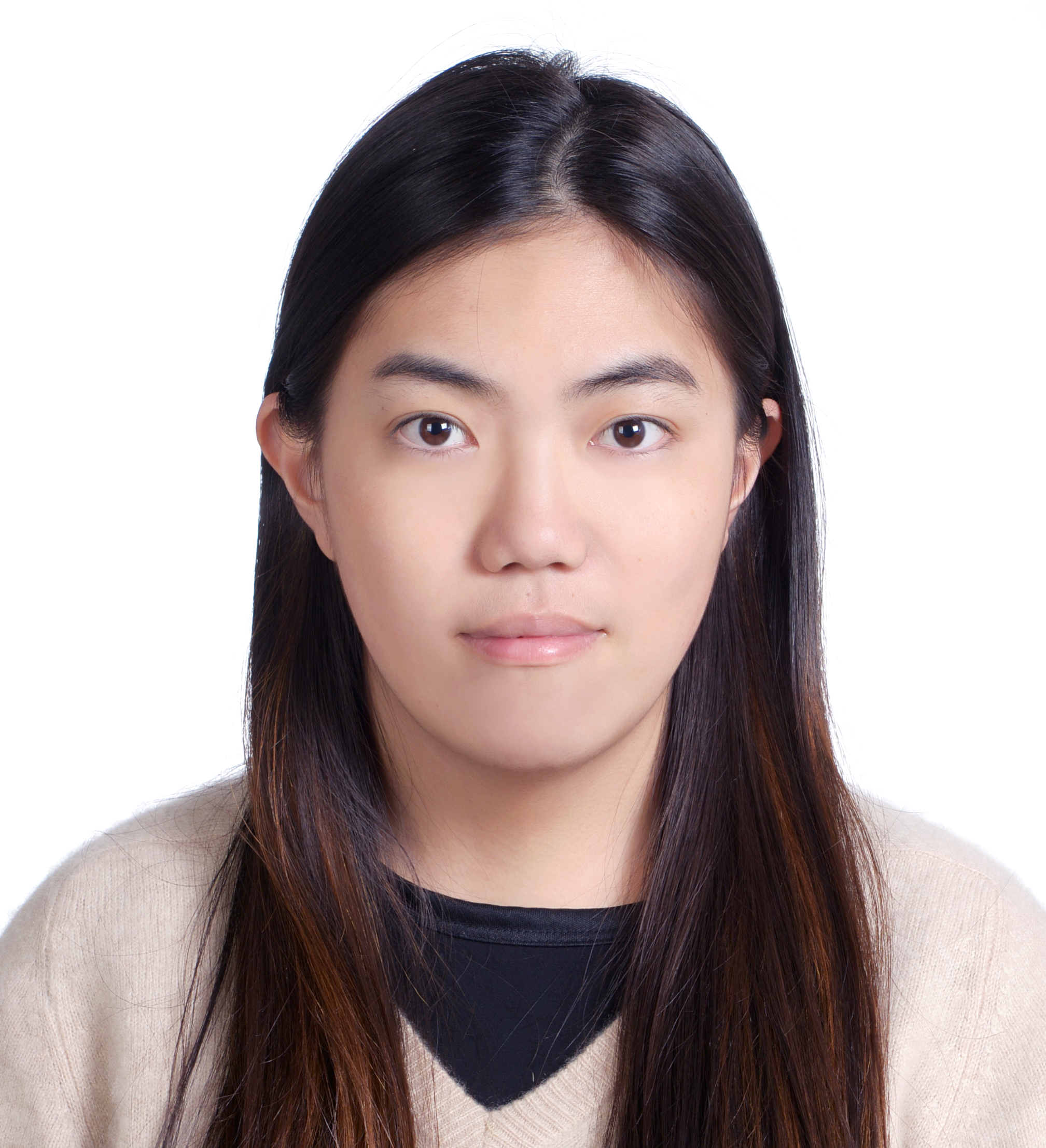}}]%
{Mei-Hsuan Wu} received the M.S. degree in Data Science from National Cheng Kung University (NCKU), Taiwan, in 2022. Her research interests include image processing, computer vision, and video DeepFake detection.
\end{IEEEbiography}

\vspace{-10mm}

\begin{IEEEbiography}[{\includegraphics[width=1in,height=1in,clip,keepaspectratio]{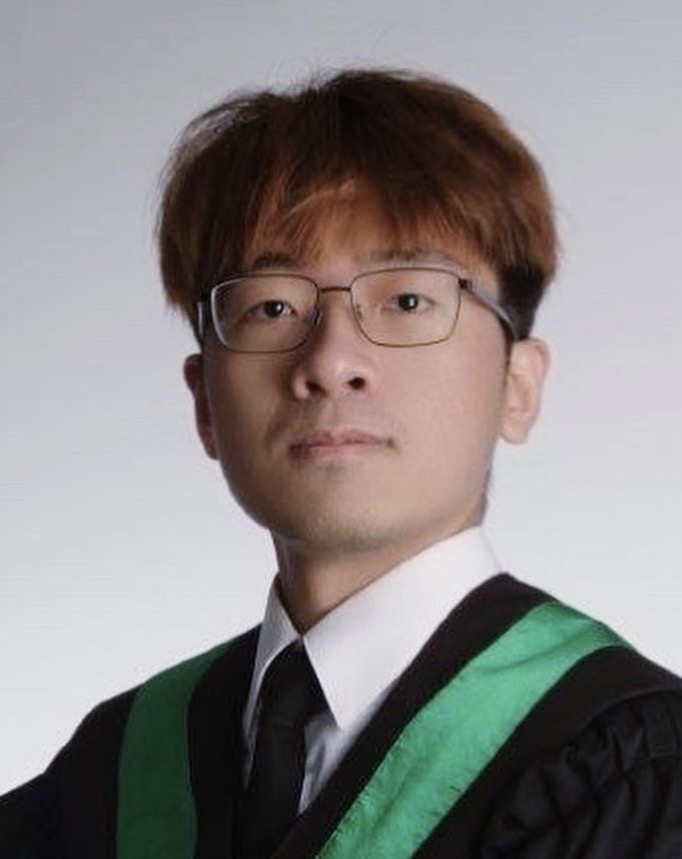}}]%
{Chia-Ming Lee} (Member, IEEE) received the B.S. degree from Fu Jen Catholic University, Taiwan, in 2023. He is currently an M.S. student with the Advanced Computer Vision Laboratory (ACVLab) at the Institute of Data Science, NCKU. His research focuses on information forensics and deep learning. He has received numerous awards, including the Jury Prize at the ICCV 2023 VIP Workshop, 1st Place in the ICASSP 2023 AI-MIA Challenge, and top honors in challenges at CVPR, ICIP, and ICPR.
\end{IEEEbiography}

\vspace{-10mm}

\begin{IEEEbiography}[{\includegraphics[width=1in,height=1in,clip,keepaspectratio]{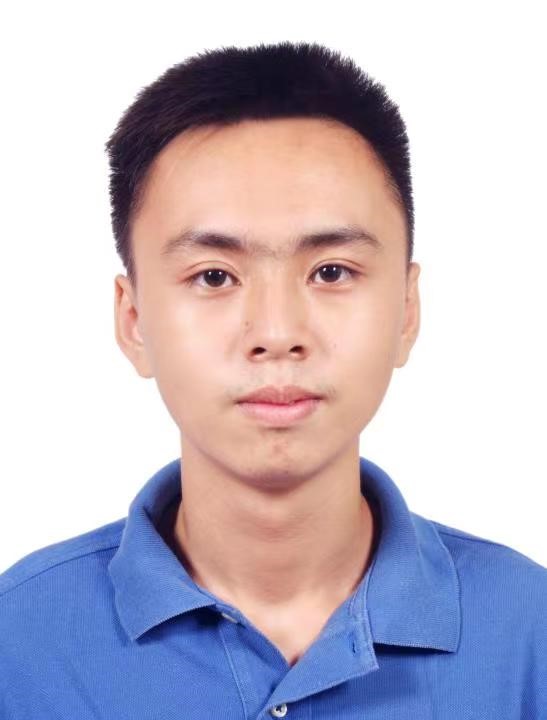}}]%
{Yi-Fan Wang} received the B.S. degree from Providence University, Taiwan, in 2022. He is currently pursuing the M.S. degree at the Institute of Data Science, National Cheng Kung University (NCKU), Taiwan. His research interests include computer vision and DeepFake detection.
\end{IEEEbiography}

\vspace{-10mm}

\begin{IEEEbiography}[{\includegraphics[width=1in,height=1in,clip,keepaspectratio]{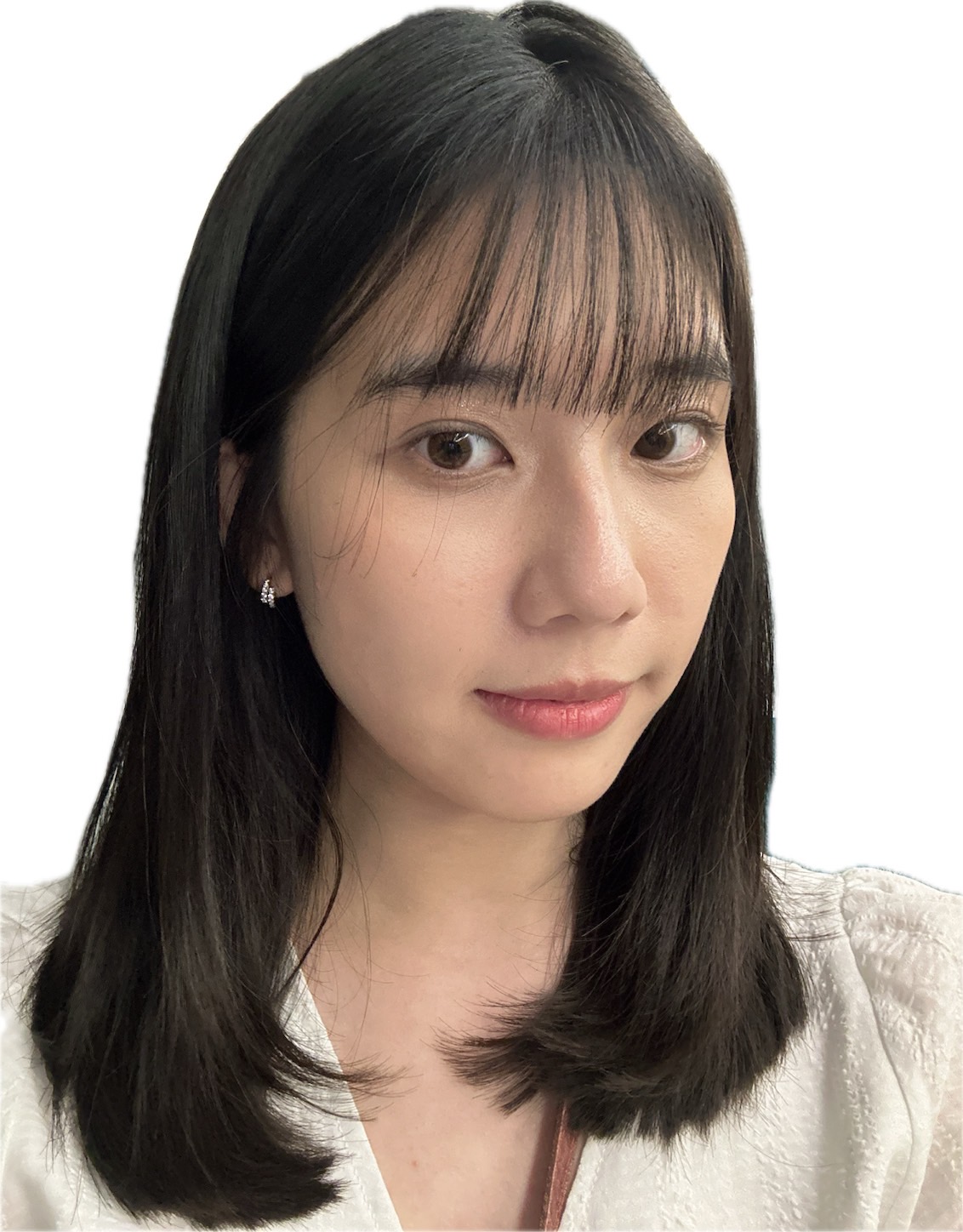}}]%
{Yi-Shiuan Chou} received the B.S. degree from National Cheng Kung University (NCKU), Taiwan, in 2024. She is currently a research assistant at the Advanced Computer Vision Laboratory (ACVLab), NCKU. Her research focuses on computer vision and deep learning. She secured 3rd place in the CVPR 2024 COVID-19 Detection Challenge and achieved top-tier performance in the NTIRE 2024 SISR Challenge and ACM Multimedia SMP Challenge.
\end{IEEEbiography}

\end{document}